\def\year{2019}\relax

\documentclass[letterpaper]{article} 
\usepackage{aaai19}  
\usepackage{times}  
\usepackage{helvet}  
\usepackage{courier}  
\usepackage{url}  
\usepackage{graphicx}  
\usepackage{amssymb,amsfonts,latexsym,color}

\usepackage{algorithm}
\usepackage{algorithmic}
\usepackage{subcaption}
\usepackage{amsmath}

\title{Message-Dropout: An Efficient Training Method for Multi-Agent Deep Reinforcement Learning}
\author{Woojun Kim \and Myungsik Cho \and Youngchul Sung \\
School of Electrical Engineering, KAIST, Korea \\
\{woojun.kim, ms.cho, ycsung\}@kaist.ac.kr}

\begin{document}

\maketitle

\begin{abstract}
In this paper, we propose a new learning technique named {\em message-dropout} to improve the performance for
 multi-agent deep reinforcement learning under two application scenarios: 1) classical multi-agent reinforcement learning with direct message communication among agents and 2) centralized training with decentralized execution. In the first application scenario of multi-agent systems in which direct message communication among agents is allowed,
the message-dropout technique drops out the received messages from other agents in a block-wise manner with a certain probability in the training phase and compensates for this effect by multiplying the weights of the dropped-out block units with a correction probability. The applied message-dropout technique effectively handles the increased input dimension in multi-agent reinforcement learning with communication and makes  learning robust against communication errors in the execution phase.
In the second application scenario of centralized training with decentralized execution, we particularly consider the application of the proposed message-dropout to Multi-Agent Deep Deterministic Policy Gradient (MADDPG), which uses a centralized critic to train a decentralized actor for each agent. We evaluate the proposed message-dropout technique for several games, and numerical results show that the proposed message-dropout technique with proper dropout rate improves the reinforcement learning performance significantly in terms of the training speed and the steady-state performance in the execution phase.
\end{abstract}

\section{Introduction}

Multi-Agent Deep Reinforcement Learning (MADRL) is gaining increasing attention from the research community with the recent success of deep learning because many of practical decision-making problems such as connected self-driving cars and collaborative drone navigation  are modeled as multi-agent systems requiring action control.
There are mainly two approaches in MADRL: one is centralized control and the other is decentralized control.
The centralized control approach assumes that there exists a central controller which determines the actions of all agents based on all the observations of all agents.
That is, the central controller has a policy which maps the joint observation to a joint action.
Since the action is based on the joint observation, this approach eases the problem of lack of full observability of the global state in partially observable environments \cite{Goldman&Zilberstein:04JAIR}.
However, this approach has the problem of the curse of dimensionality because the state-action space grows exponentially as the number of agents increases \cite{Busonui&Babuska&Schutter:08IMAS}.
Moreover, exploration, which is essential in RL, becomes more difficult than the single-agent RL case due to the huge state-action space.
Hence, to simplify the problem, the decentralized control approach was considered.
In fully decentralized multi-agent control, each agent decides its action based only on own observation, while treating other agents as a part of the environment, to reduce the curse of dimensionality. However, this approach eliminates the cooperative benefit from the presence of other agents and suffers from performance degradation.

In order to improve the performance of the decentralized control, several methods have been studied.
First, multi-agent systems with {\em decentralized control with communication (DCC)} was studied \cite{Goldman&Zilberstein:04JAIR}.
In the framework of MADRL with DCC, the agents communicate with each other by sending messages both in the training and execution phases, and the policy of each agent parameterized by a deep neural network determines the action of the agent based on own observation and the received messages from other agents.
To incorporate the messages from other agents, the size of the deep neural network of each agent should be increased as the number of message-passing agents increases. However, if the network size becomes too large, training becomes difficult and may fail.
Another approach is {\em centralized learning with decentralized execution}, allowing each agent to use the information of other agents only in the training phase.
In particular, recently-introduced MADDPG \cite{Lowe&Ryan:17NIPS}, which uses a centralized critic to train a decentralized policy for each agent, belongs to this second category.
In MADDPG, the centralized critic takes all of other agents' observations and actions as input and hence  the input space of each critic exponentially grows with the number of agents.
In both approaches, as the number of agents in the system increases, the input dimension increases,  learning becomes difficult, and much  data is required for training. Hence, it is an important problem to properly handle the increased input dimension and devise an efficient learning algorithm for such MADRL with information exchange.

In this paper,  motivated from dropout \cite{Srivastava&Hinton&Krizhevsky&Sutskever&Salakhutdinov:14JAIR}, we  propose a new training method, named {\em message-dropout}, yielding efficient learning for   MADRL with information exchange with large  input dimensions.
The proposed method improves learning performance when it is applied to MADRL with information exchange.
Furthermore, when it is applied to the scenario of DCC,  the proposed method makes learning robust against communication errors in the execution phase.

\section{Background}
\label{sec:Background}

\subsection{A Partially Observable Stochastic Game}
\label{subsec:POSG}

In MADRL, multiple agents learn how to act to maximize their future rewards while sequentially interacting with the environment.
The procedure can be described by a Partially Observable Stochastic Game (POSG) defined by the tuple $<\mathcal{I}, \mathcal{S},\{\mathcal{A}_i\},\{\Omega_i\}, \mathcal{T},\mathcal{O},\{\mathcal{R}_i\}>$, where $\mathcal{I}$ is the set of agents  $\{1,\cdots,N\}$, $\mathcal{A}_i$ is the action space of agent $i$, and $\Omega_i$ is the observation space of agent $i$.
At each time step $t$, the environment has a global state $s_t\in \mathcal{S}$ and agent $i$ observes its local observation $o_t^i \in \Omega_i$, which is determined by
the observation probability $\mathcal{O} : \mathcal{S} \times \overrightarrow{\mathcal{A}} \times \overrightarrow{
\Omega} \rightarrow [0,1]$, where $\overrightarrow{\mathcal{A}}=\prod_{i=1}^{N}\mathcal{A}_i$ and $\overrightarrow{
\Omega}=\prod_{i=1}^{N}\Omega_i$ are the joint action space and the joint observation space, respectively.
Agent $i$ executes action $a_t^i \in \mathcal{A}_i$, which yields the next global state $s_{t+1}$ with the state transition probability $\mathcal{T} : \mathcal{S} \times \overrightarrow{\mathcal{A}} \times \mathcal{S} \rightarrow [0,1]$,   receives the reward $r_t^i$ according to the reward function $\mathcal{R}_i : \mathcal{S}\times \overrightarrow{\mathcal{A}} \rightarrow {\mathbb{R}}$, and obtains the next observation $o_{t+1}^i$.
 The discounted return for agent $i$ is defined as $R_t^i = \sum_{t'=t}^{\infty} \gamma^{t'} r_{t'}^i$ where $\gamma \in [0,1]$ is the discounting factor. In POSG,
  the Q-function $Q^i(s,a^i)$ of agent $i$ can be approximated by $Q^i(\tau^i,a^i)$, where $\tau^i$ is the joint action-observation history of agent $i$. However, learning the action-value function based on the action-observation history is difficult. In this paper, we simply approximate $Q^i(s,a^i)$ with $Q^i(o^i,a^i)={\mathbb{E}}[R^i_t|o^i_t=o^i,a^i_t=a^i]$. Note that a recurrent neural network can be used to minimize the approximation error between $Q^i(s,a^i)$ and $Q^i(o^i,a^i)$ \cite{SDMIA15-Hausknecht}.
 The goal of each agent is to maximize its expected return, which is equivalent to maximizing its Q-function. (Note that in this paragraph, we explained the fully decentralized case.)

\subsection{Independent Q-Learning}
Independent Q-Learning (IQL), which is  one of the popular decentralized multi-agent RL algorithms in the fully observable case \cite{Tan&Ming:93ICML}, is a simple extension of Q-learning to multi-agent setting. Each agent estimates its own optimal Q-function, $Q^*(s,a)= \mbox{argmax}_{\pi}Q^{\pi}(s,a)$, which satisfies the Bellman optimality equation $Q^*(s,a)={\mathbb{E}}[r+\gamma\mbox{max}_{a'}Q^*(s',a')|s,a]$.  Under the assumption of full observability at each agent and fully decentralized control, Tampuu {\it et al.} combined IQL with deep Q-network (DQN), and proposed that  each agent trains its Q-function parameterized by a neural network $\theta^i$
by minimizing the loss function \cite{Tampuu&Matiisen&Kodelja&Kuzovkin&Korjus&Aru&Vicente:17PLOS}
\begin{equation}\label{eq:Loss}
L(\theta^i)=\mathbb{E}_{(s,a^i,r,s')\sim\mathcal{D}^i}\left[(y^i-Q^i(s,a^i;\theta^i))^2\right]
\end{equation}
where $\mathcal{D}^i$ is the replay memory and $y^i = r+\gamma\mbox{max}_{a^i}Q(s',a^i;\theta^{i-})$ is the target Q-value for agent $i$. Here, $\theta^{i-}$ is the target network for agent $i$.

In the case of POSG with fully decentralized control, the above loss function can be modified to
\begin{equation}\label{eq:Loss2}
L(\theta^i)={\mathbb{E}}_{(o^i,a^i,r^i,(o^i)')\sim\mathcal{D}^i}\left[(y^i-Q^i(o^i,a^i;\theta^i))^2\right]
\end{equation}
where $y^i = r^i+\gamma\mbox{max}_{a^i}Q(o^i,a^i;\theta^{i-})$. Here, $Q^i(s,a^i;\theta^i)$ in the fully observable case is approximated with $Q^i(o^i,a^i;\theta^i)$ with the local partial observation $o^i$, as described in Section \ref{subsec:POSG}.

\subsection{Multi-Agent DDPG}

 As an extension of DDPG to multi-agent setting, MADDPG was proposed to use a decentralized policy with a centralized critic for each agent \cite{Lowe&Ryan:17NIPS}.
The centralized critic uses additional information about the policies of other agents, and this helps learn the policy effectively in the training phase.
The centralized critic for agent $i$ is represented by $Q^i_{\mu}(\mathbf{x},\overrightarrow{a};\theta^i_Q)$ parameterized by a neural network $\theta^i_Q$, where $\mu=\{\mu^1, ..., \mu^N \}$ is the collection of all agents' deterministic policies,  $\mathbf{x}=(o^1,\cdots, o^N)$, and $\overrightarrow{a}=(a^1,\cdots,a^N)$. Each agent trains its Q-function by minimizing the loss function
\[
\mathcal{L}(\theta^i_Q) = {\mathbb{E}}_{\mathbf{x},a,\mathbf{x}'  \sim \mathcal{D}} \left[ (y^i - Q^i_{\mu}(\mathbf{x},\overrightarrow{a};\theta^i_Q))^2 \right]
\]
where   $y^i= r^i + \gamma Q^{i}_{\mu'}$ $(\mathbf{x}',\overrightarrow{a'};\theta^{i-}_Q)|_{{a'}^j={\mu'}^j({o'}^j)}$, $\mathbf{x}'=({o'}^1,\cdots,{o'}^N)$, and $\mathcal{D}$ is the replay memory.  Here, $\mu' = \{ {\mu'}^1 , \cdots, {\mu'}^N \}$ is the set of target policies and $\theta^{i-}_{Q}$ is the parameter of the target Q network for agent $i$.

Then, the policy for agent $i$, which is parameterized by $\theta_{\mu}^i$, is trained by deterministic policy gradient  to maximize the objective $J(\theta_{\mu}^i) = \mathbb{E}\left[\mathcal{R}_i \right]$, and the gradient of the objective is given by
\[
\nabla_{\theta_{\mu}^i} J(\mu^i) = {\mathbb{E}}_{\mathbf{x},a \sim \mathcal{D}} \left[ \nabla_{\theta_{\mu}^i}\mu^i(o^i) \nabla_{a^i} Q_{\mu}^i (\mathbf{x}, \overrightarrow{a}))|_{a^i = \mu^i(o^i)} \right]
\]

\subsection{Dropout}

Dropout is a successful neural network technique.  For a given neural network, constituting units (or nodes) in the neural network are randomly dropped out  with probability $p$ independently in the training phase to avoid co-adaptation among the units in the neural network \cite{Srivastava&Hinton&Krizhevsky&Sutskever&Salakhutdinov:14JAIR}. In the test phase, on the other hand,
all the units are included but the outgoing weights of those units affected by dropout are multiplied by $1-p$.
There is a good interpretation of dropout: efficient model averaging over randomly generated neural networks in training.
For a neural network with $N$ units, dropout samples the network to be trained from $2^N$ differently-thinned networks which share the parameter in the training phase due to independent dropping out of the $N$ units.
Scaling the weights of the units at the test phase can be regarded as averaging over the ensemble of subnetworks. It is shown that in the particular case of a single logistic function, dropout corresponds to geometric averaging \cite{Baldi&Sadowski:13NIPS}. Thus, dropout efficiently combines the exponentially-many different neural networks as one network and can also be regarded as a kind of ensemble learning \cite{Hara&Saitoh&Shouno:16ICANN}.

\begin{figure}
  \centering
  \begin{subfigure}[t]{0.2\textwidth}
    \centering
    \includegraphics[scale=0.15]{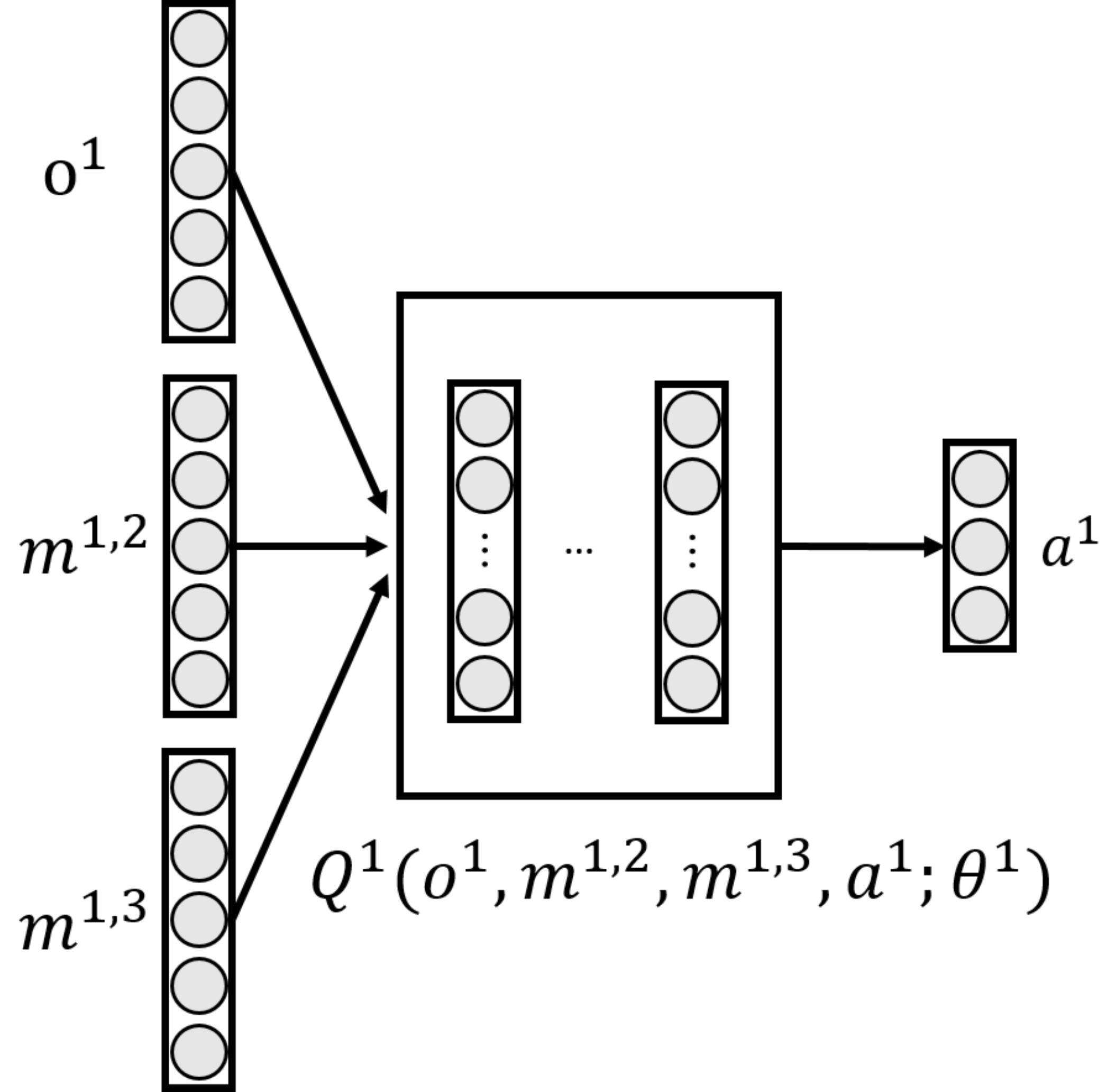}
    \caption{}
    \label{Q1}
  \end{subfigure}
  \begin{subfigure}[t]{0.2\textwidth}
    \centering
    \includegraphics[scale=0.15]{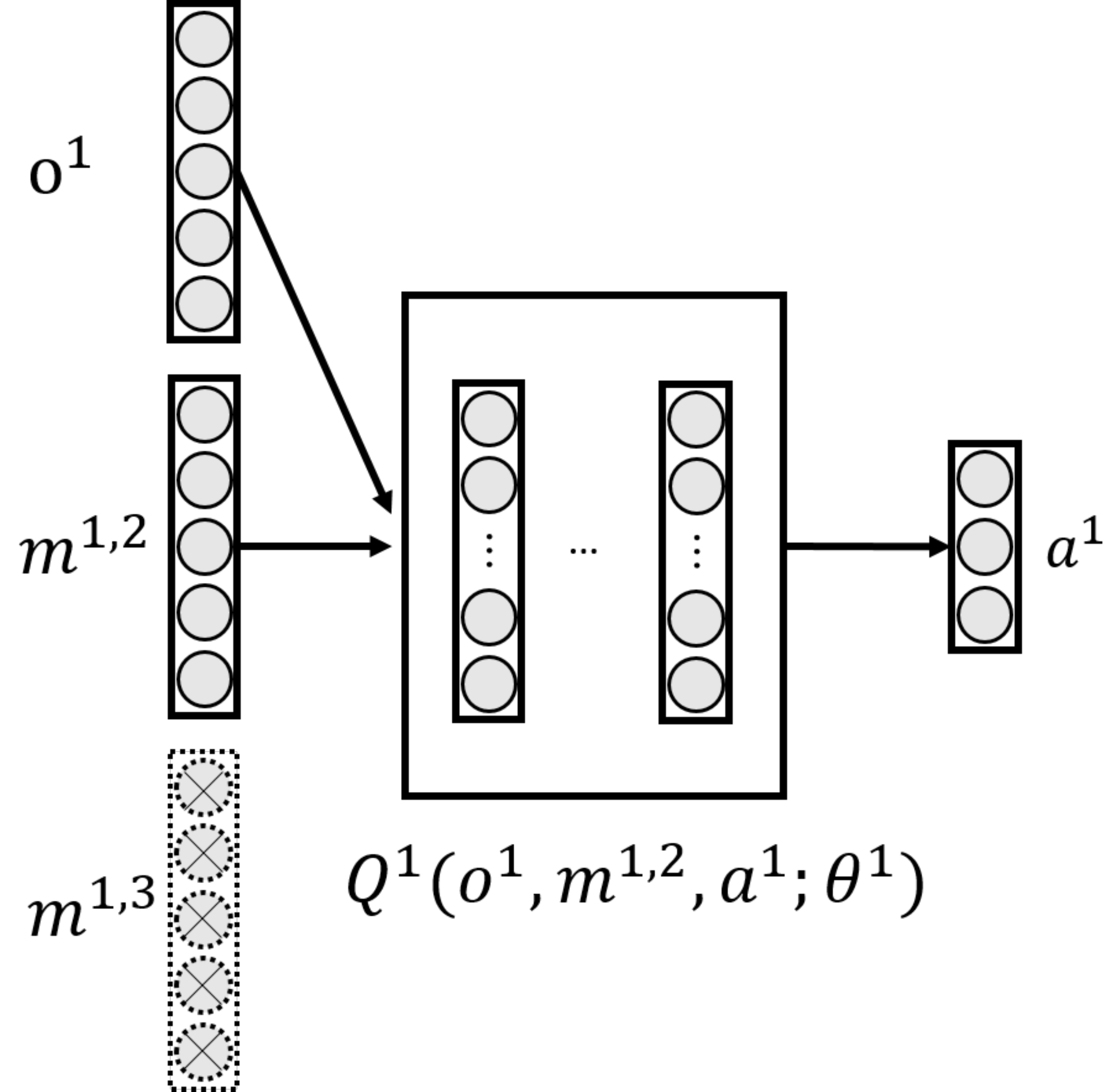}
    \caption{}
    \label{Q2}
  \end{subfigure}
  \begin{subfigure}[t]{0.2\textwidth}
    \centering
    \includegraphics[scale=0.15]{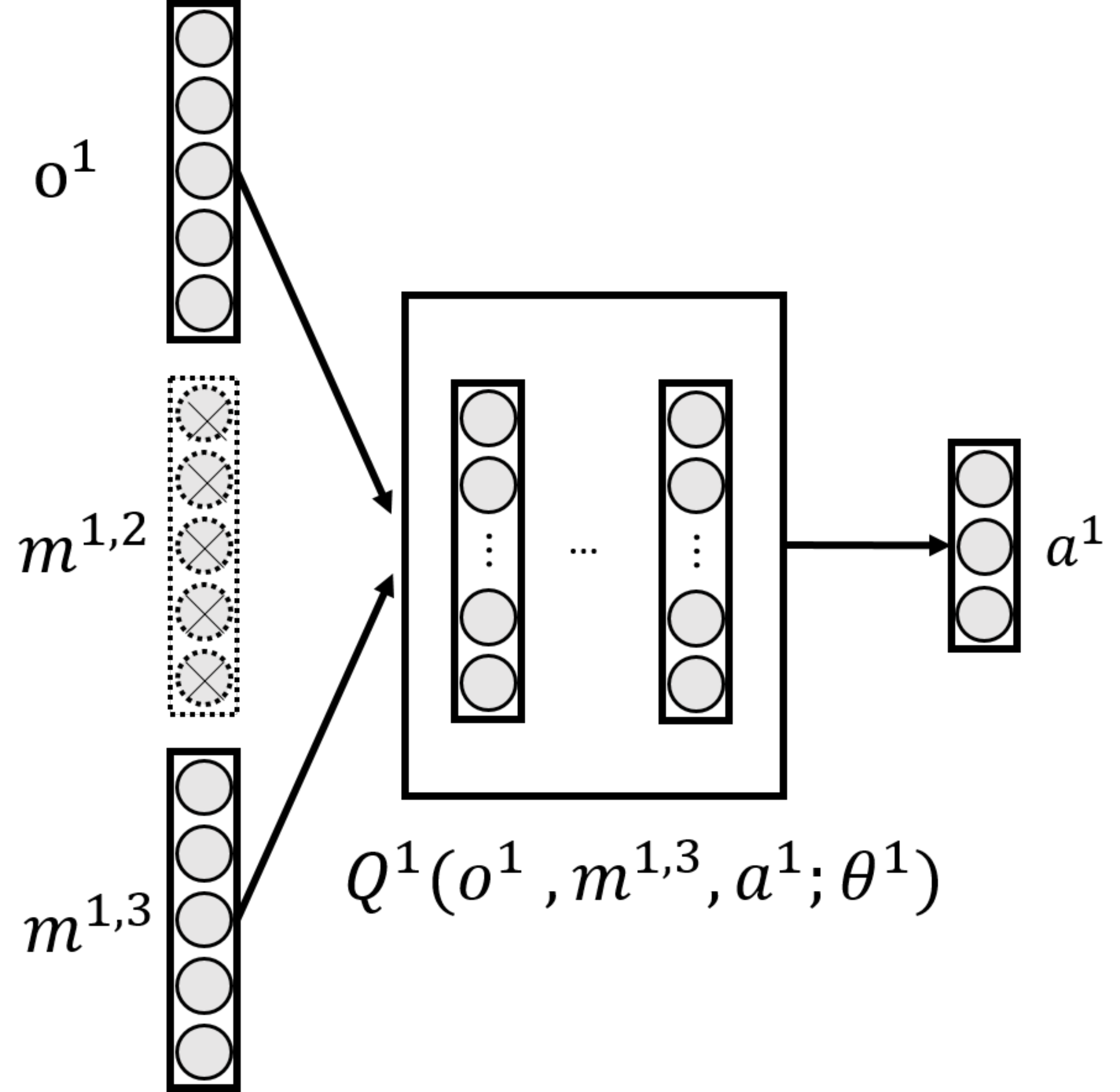}
    \caption{}
    \label{Q3}
  \end{subfigure}
  \begin{subfigure}[t]{0.2\textwidth}
    \centering
    \includegraphics[scale=0.15]{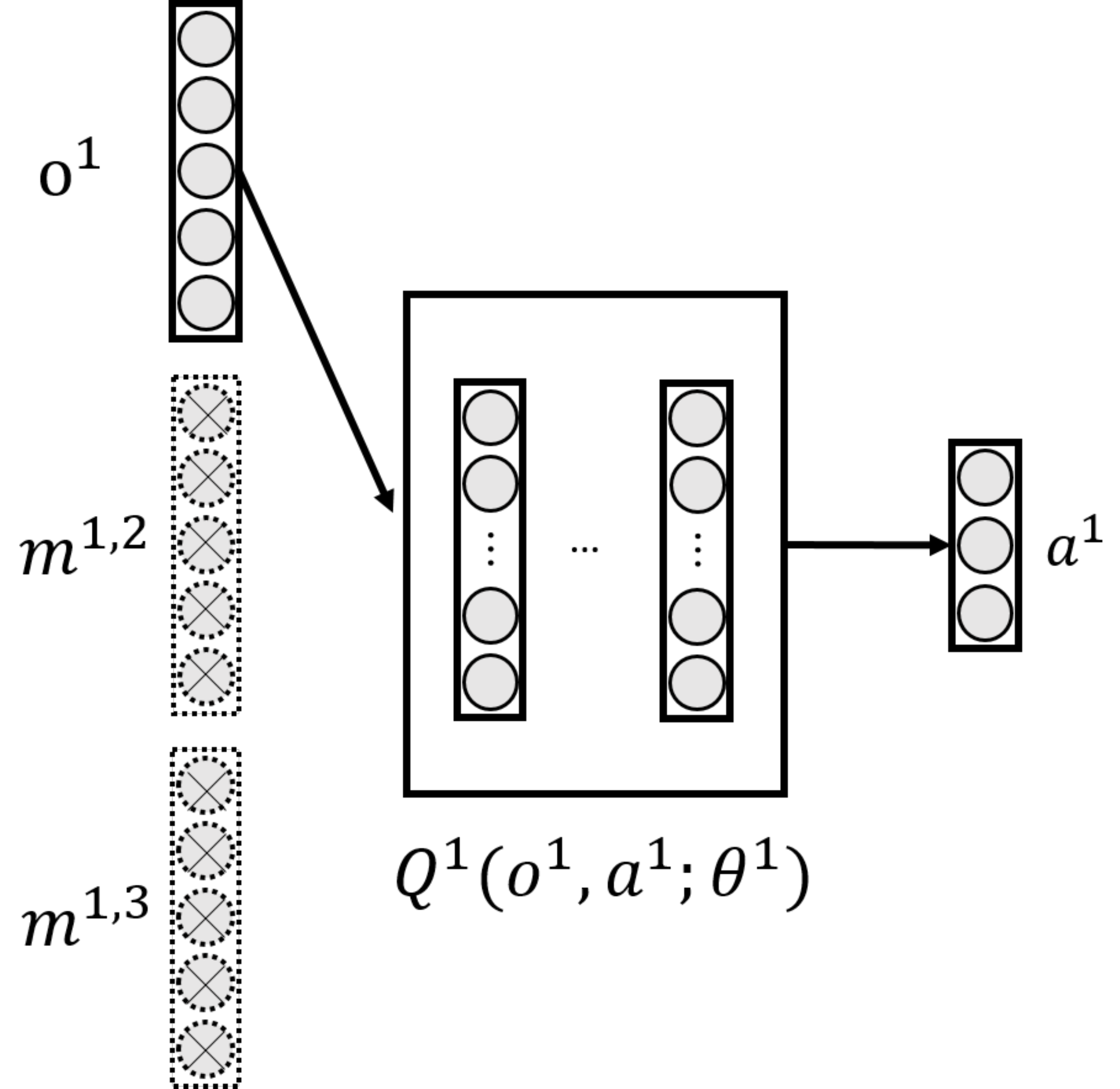}
    \caption{}
    \label{Q4}
  \end{subfigure}
  \label{4candidate}
  \caption{$N=3$: All candidates for agent 1's Q-network}
  \label{fig:4candidate}
\end{figure}

\section{Methodology}
\label{Method}

\subsection{Decentralized Control with Communication}
\label{subsec:BasicSetup}

In this subsection, we consider the direct communication framework in \cite{Goldman&Zilberstein:04JAIR} for DCC. This framework can be modeled by adding messages to POSG. Each agent exchanges messages before action, and chooses action based on its own observation and the received messages. Hence, the policy of agent $i$ is given by  $\pi^i : \Omega_i \times \mathcal{M}_{i} \times \mathcal{A}_i \rightarrow [0,1]$, where  $\mathcal{M}_{i}=\prod_{j\neq i}\mathcal{M}_i^j$ and $\mathcal{M}_i^j$ is the space of received messages from agent $j$ to agent $i$. We denote agent $i$'s received message from agent $j$ as $m^{i,j} \in \mathcal{M}_i^j$, and $m^i = (m^{i,1},\cdots,m^{i,i-1},m^{i,i+1},\cdots,m^{i,N})$ is the collection of the received messages at agent $i$.
For the learning engine at each agent, we use the Double Deep Q-Network (DDQN) which alleviates the overestimation problem of Q-learning \cite{Hasselt&Guez&Silver:16AAAI}.
In this case, the Q-function of agent $i$ parameterized by the neural network $\theta^i$ is given by
 $Q^i(o^i,m^i, a^i ; \theta^i) = {\mathbb{E}}[R^i|o^i, m^i, a^i;\theta^i]$. Then, the Q-function of  agent $i$ is updated by minimizing the loss function
\begin{equation}
\label{eq:Loss3}
\resizebox{0.45\textwidth}{!}{$
L(\theta^i)=\mathbb{E}_{(o^i,m^i,a^i,r^i,{o'}^i,{m'}^i)\sim\mathcal{D}^i}\left[(y^i-Q^i(o^i,m^i,a^i;\theta^i))^2\right]
$}
\end{equation}
where \resizebox{0.43\textwidth}{!}{$y^i=r^i+\gamma Q^i({o'}^i,{m'}^i,\mbox{argmax}_{a'}Q^i({o'}^i,{m'}^i,a';\theta^i);\theta^{i-})$}.
We will refer to this scheme as {\em simple DCC}.

The problem of simple DCC is that the input dimension of the $Q$ neural network at each agent linearly increases and the size of the input state space increases exponentially,
as the number of agents increases. Thus, the number of required samples for learning increases, and this decreases the speed of learning.
Another issue of  simple DCC is that the portion of each agent's own observation space in the input space of the Q-function  decreases as the number of agents increases.
Typically, the own observation of each agent is most important. Hence, the importance of each agent's own observation is not properly weighted in simple DCC.

\subsection{Message-Dropout}
\label{block-dropout}

To address the aforementioned issues of  simple DCC, we  propose a new neural network technique, named {\em message-dropout}, which can be applied  to decentralized control with communication of messages.

Message-dropout drops out the received other agents' messages at the input of the Q-network at each agent independently in the training phase.
 That is, all units corresponding to the message from one dropped other agent are dropped out simultaneously in a blockwise fashion with probability $p$ and this blockwise dropout is independently performed across all input unit blocks corresponding to all other agents' messages. On the other hand, the outgoing weights of those input units on which dropout is applied are multiplied by $1-p$, when the policy generates actual action. Note that dropout is not applied to the input units corresponding to each agent's own observation. To illustrate, let us consider agent  $1$ in an environment with total three agents, as shown in Fig. \ref{fig:4candidate}. The Q-network of agent 1 has three input blocks: one for own observation and two for the messages from two other agents. By applying the proposed message dropout, as shown in Fig. \ref{fig:4candidate}, we have the four possible configurations for the input of the Q-network of agent 1:
\[
\resizebox{0.45\textwidth}{!}{$
(o^1,m^{1,2},m^{1,3}), (o^1,m^{1,2},\overrightarrow{0_3}), (o^1,\overrightarrow{0_2},m^{1,3}),  (o^1,\overrightarrow{0_2},\overrightarrow{0_3})
$}
\]
where $\overrightarrow{0_j} \in R^{|\mathcal{M}_i^j|}$ with all zero elements represents the dropped-out input units.

\begin{algorithm}[!t]
\caption{DCC with Message-Dropout (DCC-MD)}\label{algorithm:DCC-MD}

\begin{algorithmic}


\STATE{Initialize $\theta^1,\cdots,\theta^N$ and ${\theta^{-}}^1,\cdots,{\theta^{-}}^N$.}

\FOR {episode = $1,2,\cdots$}

\STATE{Initialize state $s_1$.}

\FOR {$t<T$ and $s_t\neq$ terminal}

\STATE{Each agent $i$ observes $o_t^i$ and sends $m_t^{i}$}

\FOR{each agent $i=1,2,\cdots,N$}

\STATE{Receive messages $m_t^{i}$.}

\STATE{With probability $\epsilon$ select a random action $a_t^i$}

\STATE{otherwise select the action $a_t^i$ from (\ref{eq:policy}).}

\ENDFOR

\STATE{Execute $\overrightarrow{a}$ and each agent $i$ receives  $r_t^i$ and $o_{t+1}^i$.}

\STATE{Each agent sends  the message to other agents.}

\FOR{each agent $i=1,2,\cdots,N$}

\STATE{Store transition \resizebox{0.22\textwidth}{!}{$(o_t^i,m_t^{i},a_t^i,r_t^i,{o}_{t+1}^i,{m}_{t+1}^{i})$} in $\mathcal{D}^i$}

\STATE{Sample $\{(o_j^i,m_j^{i},a_j^i,r_j^i,{o}_{j+1}^i,{m}_{j+1}^{i}), j\in \mathcal{J}^i\}$ from $\mathcal{D}^i$}

\STATE{Generate a binary mask and obtain $\tilde{m}_j^{i}$ and $\tilde{m}_{j+1}^{i}$}

\STATE{Update $\theta^i$ by minimizing the loss function (\ref{eq:5})}

\ENDFOR

\STATE{Update the target network}

\ENDFOR

\ENDFOR

\end{algorithmic}

\end{algorithm}

Now, we explain the Q-learning process for DCC with message-dropout (DCC-MD). Consider the learning at agent $i$. Agent $i$ stores
the transition $(o^i,m^{i},a^i,r^i,{o'}^i,{m'}^i)$ into its replay memory $\mathcal{D}^i$. To train the Q-function, agent $i$ samples a random mini-batch of transitions from $\mathcal{D}^i$, denoted $\{(o^i_j,m^i_j,a^i_j,r^i_j,{o}^i_{j+1},m^i_{j+1}), j \in \mathcal{J}^i\}$. Message-dropout is performed independently for each $j \in \mathcal{J}^i$ and the message-dropout performed observation and its transition are given by
\[
    \resizebox{0.47\textwidth}{!}{$ \begin{split}
    \tilde{m_j}^{i}&=(b_{j,1} m_j^{i,1},\cdots,b_{j,i-1} m_j^{i,i-1},b_{j,i+1} m_j^{i,i+1},\cdots,b_{j,N} m_j^{i,N})  \\
    \tilde{m}_{j+1}^i &= (b_{j,1} m_{j+1}^{i,1},\cdots,b_{j,i-1} m_{j+1}^{i,i-1},b_{j,i+1} m_{j+1}^{i,i+1},\cdots,b_{j,N} m_{j+1}^{i,N})
    \end{split}$}
\]
where the scalar value $b_{j,k}$ $\sim$ Bernoulli($p$). Note that the same binary mask is used to define $\tilde{m}_j^{i}$ and $\tilde{m}_{j+1}^{i}$. Then, the Q-function is updated by minimizing the loss
\begin{equation} \label{eq:5}
\resizebox{0.45\textwidth}{!}{$
L(\theta^i) = \mathbb{E}_{(o_j^i,m_j^{i},a_j^i,r_j^i,o_{j+1}^i,m_{j+1}^{i})\sim\mathcal{D}^i}\left[(y^i_j-Q^i(o_j^i,\tilde{m_j}^{-i},a^i_j;\theta^i))^2\right]
$}
\end{equation}
where \resizebox{0.43\textwidth}{!}{$y^i_j = r^i_j+\gamma Q^i(o_{j+1}^i,\tilde{m_{j+1}}^i,\mbox{argmax}_{a^i}Q^i_{exec}(o_{j+1}^i,m_{j+1}^i,a^i;\theta^i);\theta^{i-})$}
Here,  $Q_{exec}^i(o^i,m^i,a^i;\theta^i)$ is the Q-function parameterized by the neural network whose outgoing weights of $m^i$ are multiplied by $1-p$. Note that we use $Q^i_{exec}$ to predict the next action when evaluating the target Q-value $y_j^i$. Finally, the policy $\pi^i$ is given by
\begin{equation}
\label{eq:policy}
\pi^i = \mbox{argmax}_{a^i} Q_{exec}^i(o^i,m^{i},a^i;\theta^i).
\end{equation}

\subsubsection{Interpretation}
\label{subsec:interpretation}

In the training phase with message-dropout, agent $i$ drops out some of other agents' messages in $m^{i}$, while keeping own observation.
As a result, the input space of the Q-network of agent $i$ is projected onto a different subspace (of the original full input space) that always includes own observation space at each training time since the dropout masks change at each training time.
The input spaces of the four Q-networks in the example of Fig. \ref{fig:4candidate} are $\Omega_1 \times \mathcal{M}_1^2 \times \mathcal{M}_1^3$, $\Omega_1 \times \mathcal{M}_1^2$, $\Omega_1 \times \mathcal{M}_1^3$, $\Omega_1$ (all include agent's own observation $\Omega_i$).
In the general case of $N$ agents, message-dropout samples the network to be trained from $2^{N-1}$ differently-thinned networks which always include the agent's own observation.
Note that the Q-network whose all received messages are retained is the Q-network of simple DCC and the Q-network whose all observations from other agents are dropped is the Q-network for fully decentralized DDQN without communication.
Thus, $2^{N-1}$ differently-thinned networks include the Q-networks of simple DCC and fully decentralized DDQN. Message-dropping out in the training phase and multiplying the outgoing weights of $m^{i}$ by $1-p$ in the test phase yields an approximate averaging over the ensemble of those networks.
Note that simple DCC and fully decentralized DDQN are the special cases of the dropout rate $p=0$ and $p=1$, respectively. Thus, for $0<p<1$, the proposed scheme realizes some network between these two extremes.

\subsection{MADDPG-MD}

Message-dropout  can also be applied to the framework of centralized training with decentralized execution, particularly to the setting in which each agent uses additional information from other agents during training as the input of the network.
For example, in MADDPG, the centralized critic $Q^i_{\mu}(\mathbf{x},\overrightarrow{a})$ takes all agents' observations and actions as input, and hence the input space of the centralized critic for each agent increases exponentially as the number of agents increases. The proposed technique, message-dropout, is applied to the training phase of  the centralized critic
to address the aforementioned problem. The centralized critic with message-dropout applied is trained to minimize the loss function:
\begin{equation}
\mathcal{L}(\theta^i_Q) = \mathbb{E}_{\tilde{\mathbf{x}},a,\tilde{\mathbf{x}}'  \sim \mathcal{D}} \left[ (y^i - Q^i_{\mu}(\tilde{\mathbf{x}},\overrightarrow{a};\theta^i_Q))^2 \right],
\end{equation}
where $y^i= r^i + \gamma  Q^{i}_{\mu'}$ $(\tilde{\mathbf{x}}',\overrightarrow{a'};\theta^{i-}_Q)|_{{a'}^j={\mu'}^j({o'}^j)}$ and $\tilde{\mathbf{x}}=(o^i, \tilde{o}^{-i})$.
Note that $\tilde{o}^{-i} = (b_1 o^{1},\cdots,b_{i-1} o^{i-1},b_{i+1} o^{i+1},\cdots,b_{N} o^{N})$, where $b_{i}$ $\sim$ Bernoulli($p$), and the same binary mask is used to obtain  $\tilde{\mathbf{x}}$ and $\tilde{\mathbf{x}}'$ as in DCC-MD.
Then, the policy for agent $i$ is trained by maximizing  the objective $J(\theta_{\mu}^i) = \mathbb{E}\left[\mathcal{R}_i \right]$, and the gradient of the objective can be written as
\[
\nabla_{\theta_{\mu}^i} J(\mu^i) = \mathbb{E}_{\tilde{\mathbf{x}},a \sim \mathcal{D}} \left[ \nabla_{\theta_{\mu}^i}\mu^i(o^i) \nabla_{a^i} Q_{\mu}^i (\tilde{\mathbf{x}}, \overrightarrow{a}))|_{a^i = \mu^i(o^i)} \right].
\]
We refer to MADDPG with message-dropout applied as MADDPG-MD.

\section{Experiment}
\label{sec:Experiment}

In this section, we provide some numerical results to evaluate the proposed algorithm in the aforementioned two scenarios for MADRL with information exchange.
First, we compare DCC-MD with simple DCC and Fully Decentralized Control (FDC) in two environments of  pursuit and cooperative navigation.
Second, we compare MADDPG-MD with MADDPG and independent DDPG (simply DDPG) in the  environment of  waterworld.
Then, we provide in-depth ablation studies to understand the behavior of message-dropout depending on various parameters. Finally, we investigate  DCC-MD in  unstable environments in which some links of communication between agents can be  broken in the execution phase

Although some compression may be applied, for simplicity we here assume that each agent's message is its observation itself, which is shown to be optimal when the communication cost is ignored  in the framework of DCC \cite{Goldman&Zilberstein:04JAIR}.
Hence, $\mathcal{M}_i^j = \Omega_j$ for all agent $i$, and the policy function becomes $\pi^i : \overrightarrow{\Omega} \times \mathcal{A}_i \rightarrow [0,1]$, where $\overrightarrow{\Omega}=(\Omega_1,\cdots,\Omega_N)$. (A brief study on message-dropout with message generation based on auto-encoder applied to raw observation is given in the supplementary file of this paper. It is seen that similar performance improvement is achieved by message-dropout in the case of compressed message. Hence, message-dropout can be applied on top of message compression for MADRL with message communication.)

\begin{figure}[t]
  \centering
  \begin{subfigure}[t]{0.15\textwidth}
    \centering
    \includegraphics[height=2.2cm]{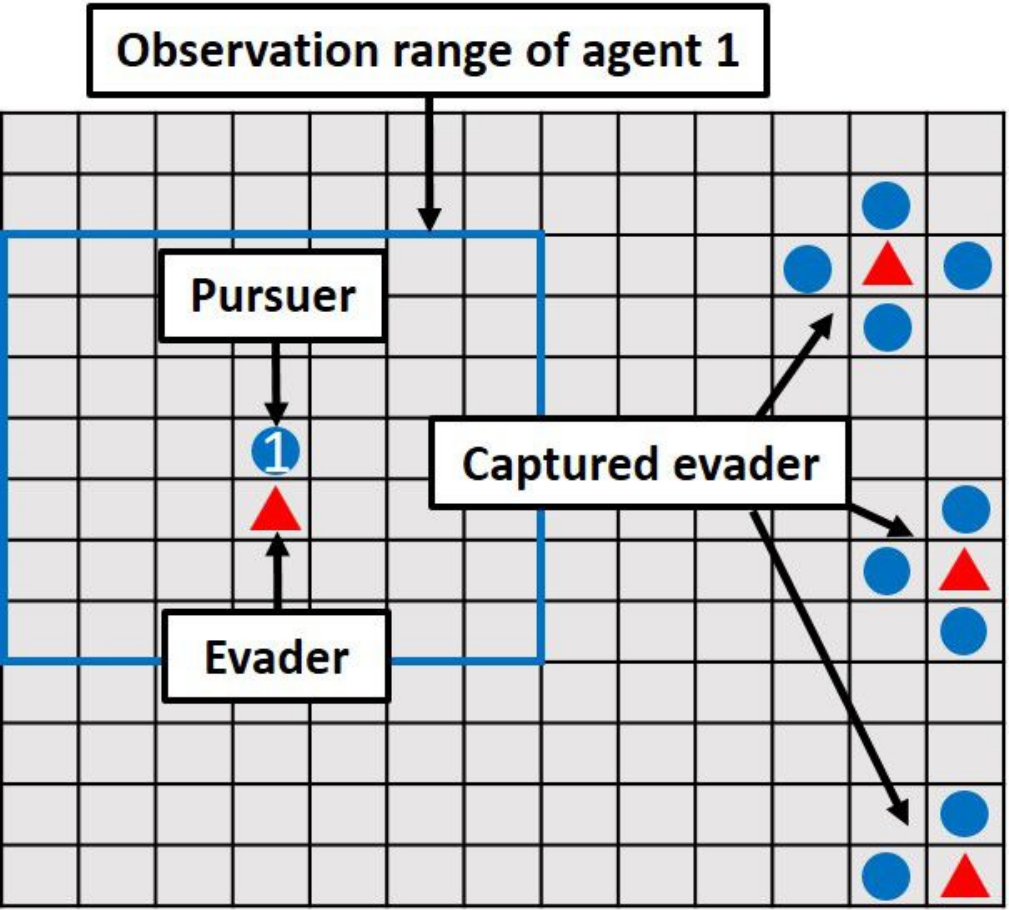}
    \caption{}
    \label{fig:env1}
  \end{subfigure}
  \begin{subfigure}[t]{0.15\textwidth}
    \centering
    \includegraphics[height=2.2cm]{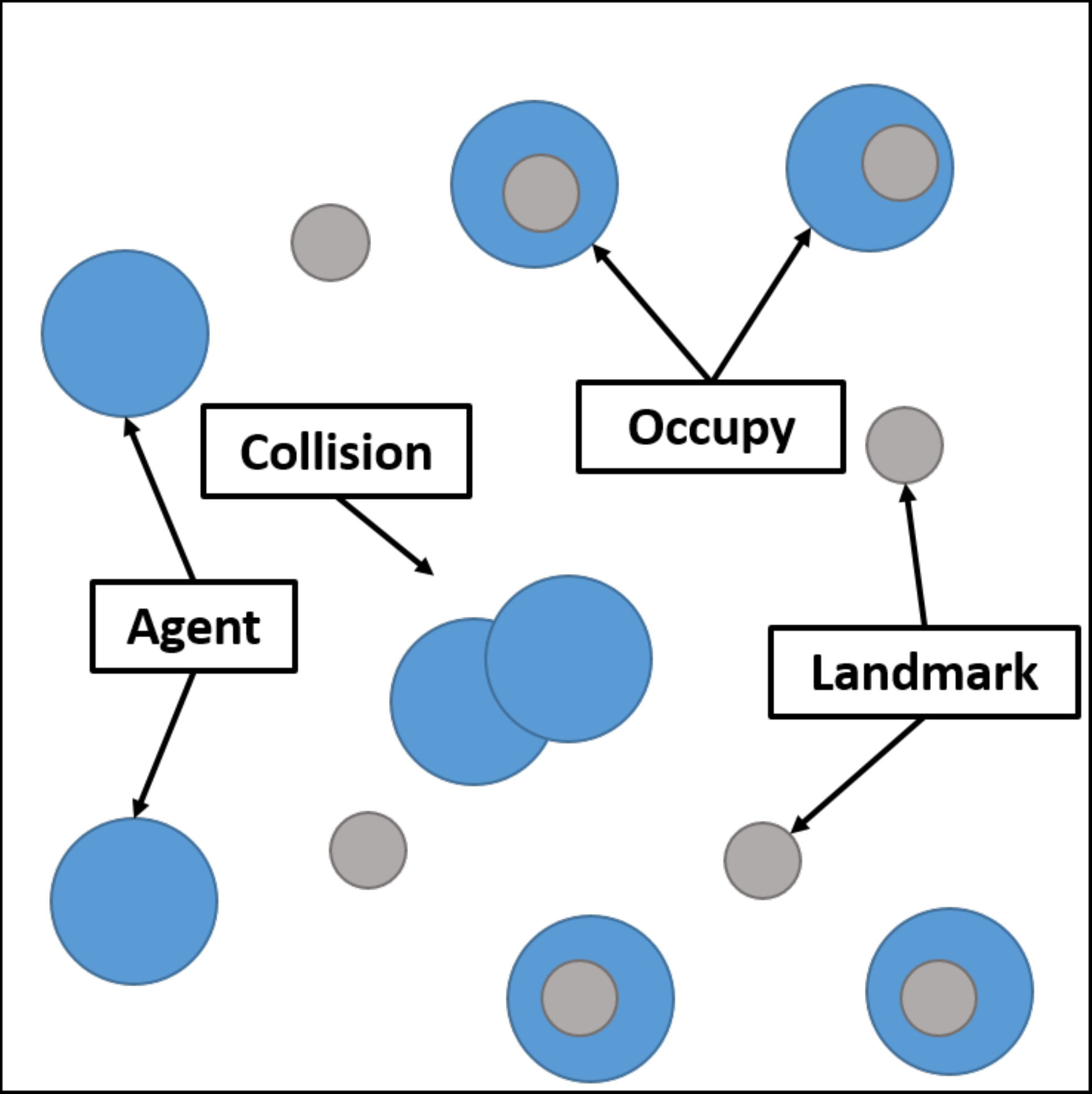}
    \caption{}
    \label{fig:env2}
  \end{subfigure}
  \begin{subfigure}[t]{0.15\textwidth}
    \centering
    \includegraphics[height=2.2cm]{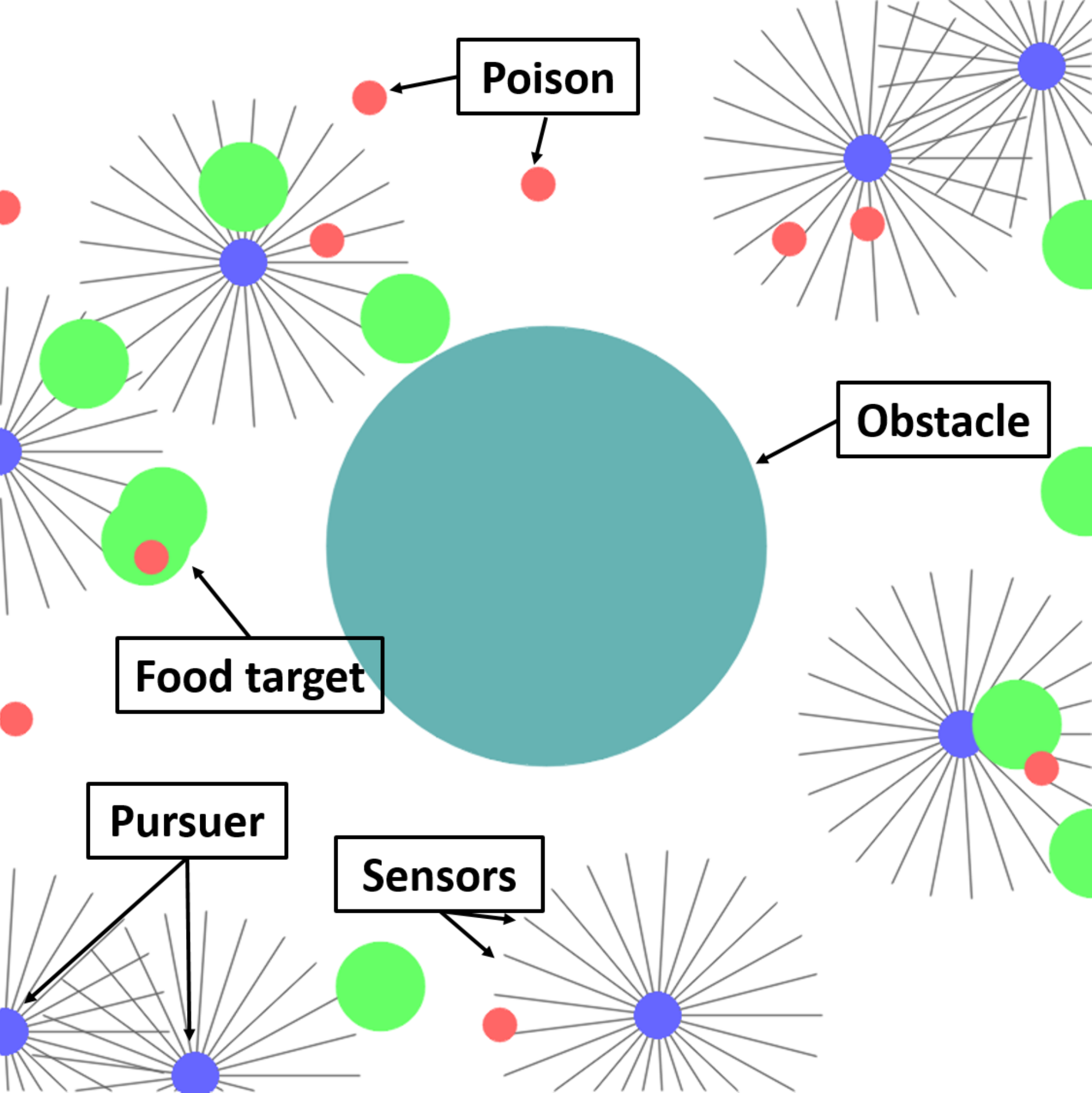}
    \caption{}
    \label{fig:env3}
  \end{subfigure}
  \caption{Considered environments: (a)  Pursuit, (b)  Cooperative navigation, and (c)  Waterworld}
  \label{fig:env}
\end{figure}

\subsection{Environment}
\label{subsec:environment}

\begin{figure*}[t]
  \centering
  \begin{subfigure}[h]{0.16\textwidth}
    \centering
    \includegraphics[height=2.3cm]{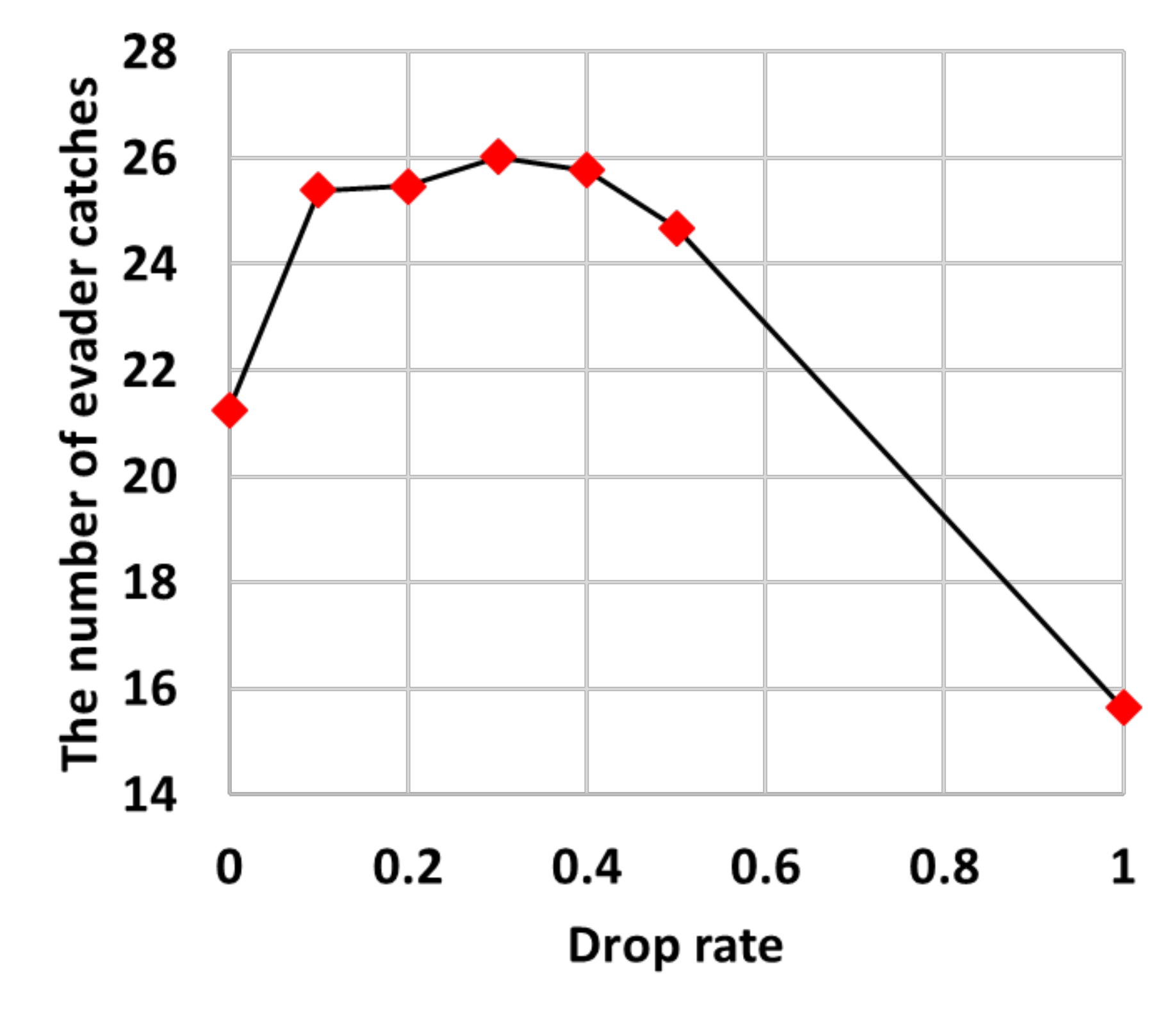}
    \caption{$N=6$}
    \label{fig:res_pur_6}
  \end{subfigure}
  \begin{subfigure}[h]{0.16\textwidth}
    \centering
    \includegraphics[height=2.3cm]{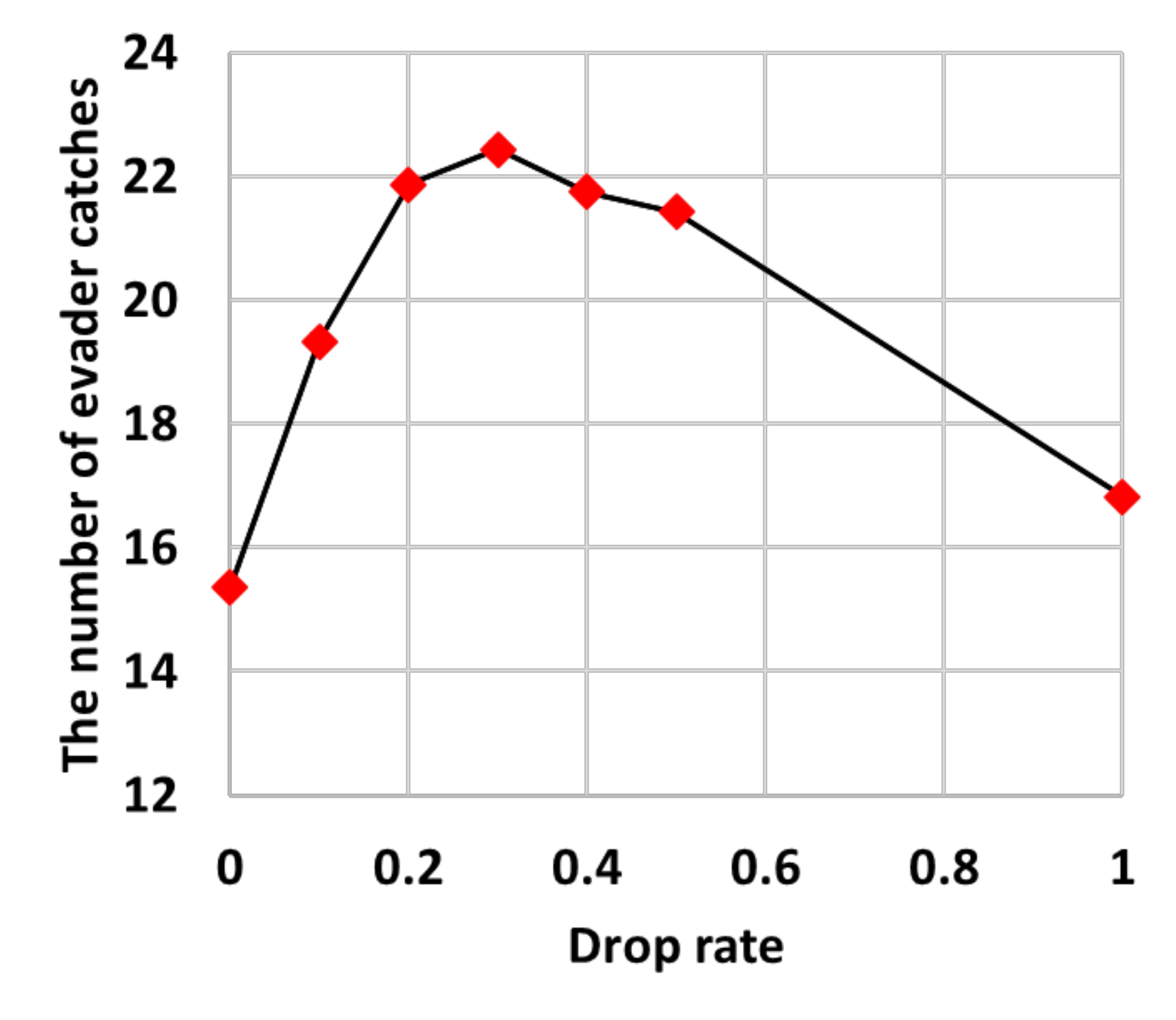}
    \caption{$N=8$}
    \label{fig:res_pur_8}
  \end{subfigure}
  \begin{subfigure}[h]{0.16\textwidth}
    \centering
    \includegraphics[height=2.3cm]{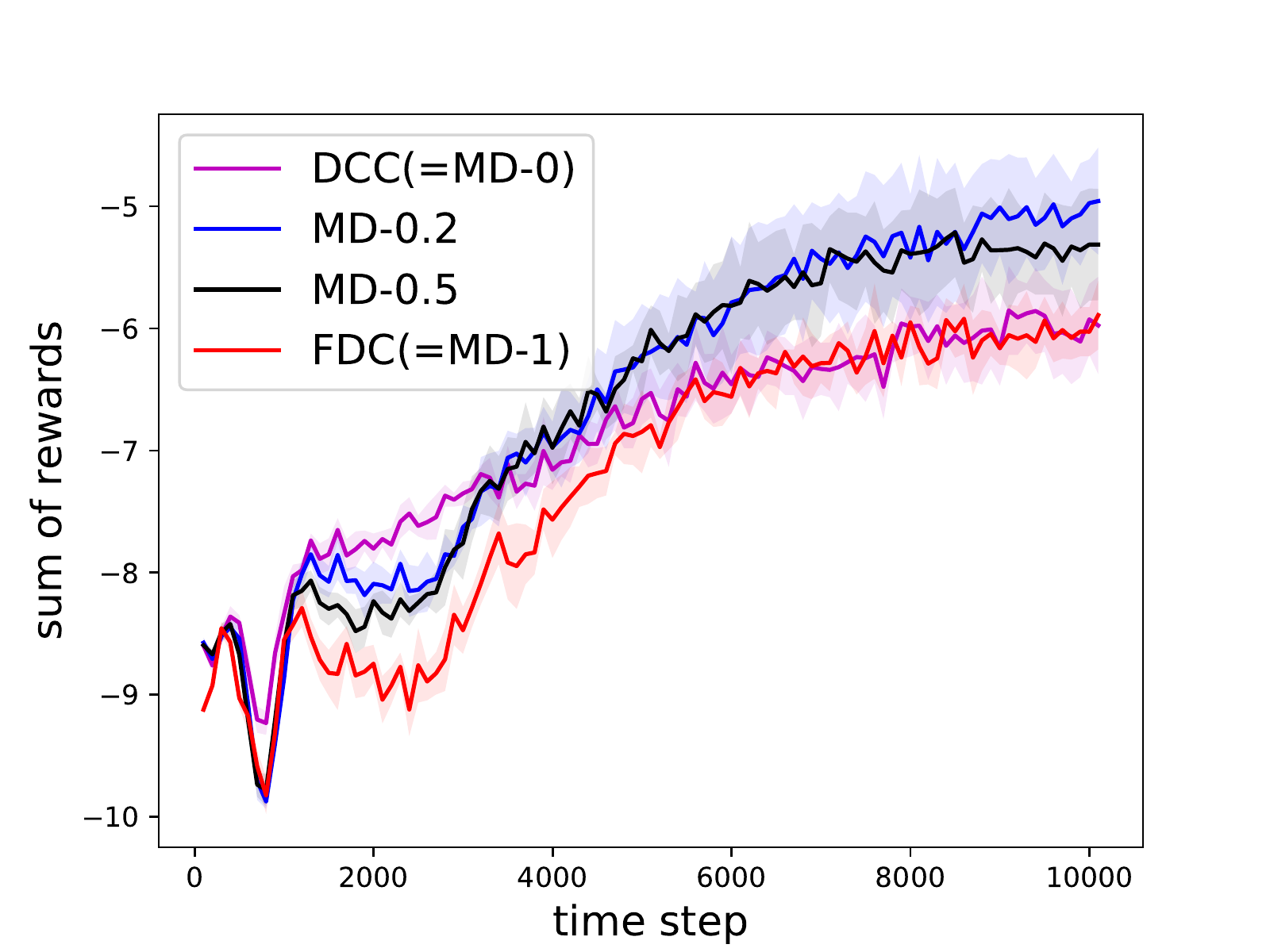}
    \caption{$N=8$}
    \label{fig:res_navi_8}
  \end{subfigure}
  \begin{subfigure}[h]{0.16\textwidth}
    \centering
    \includegraphics[height=2.3cm]{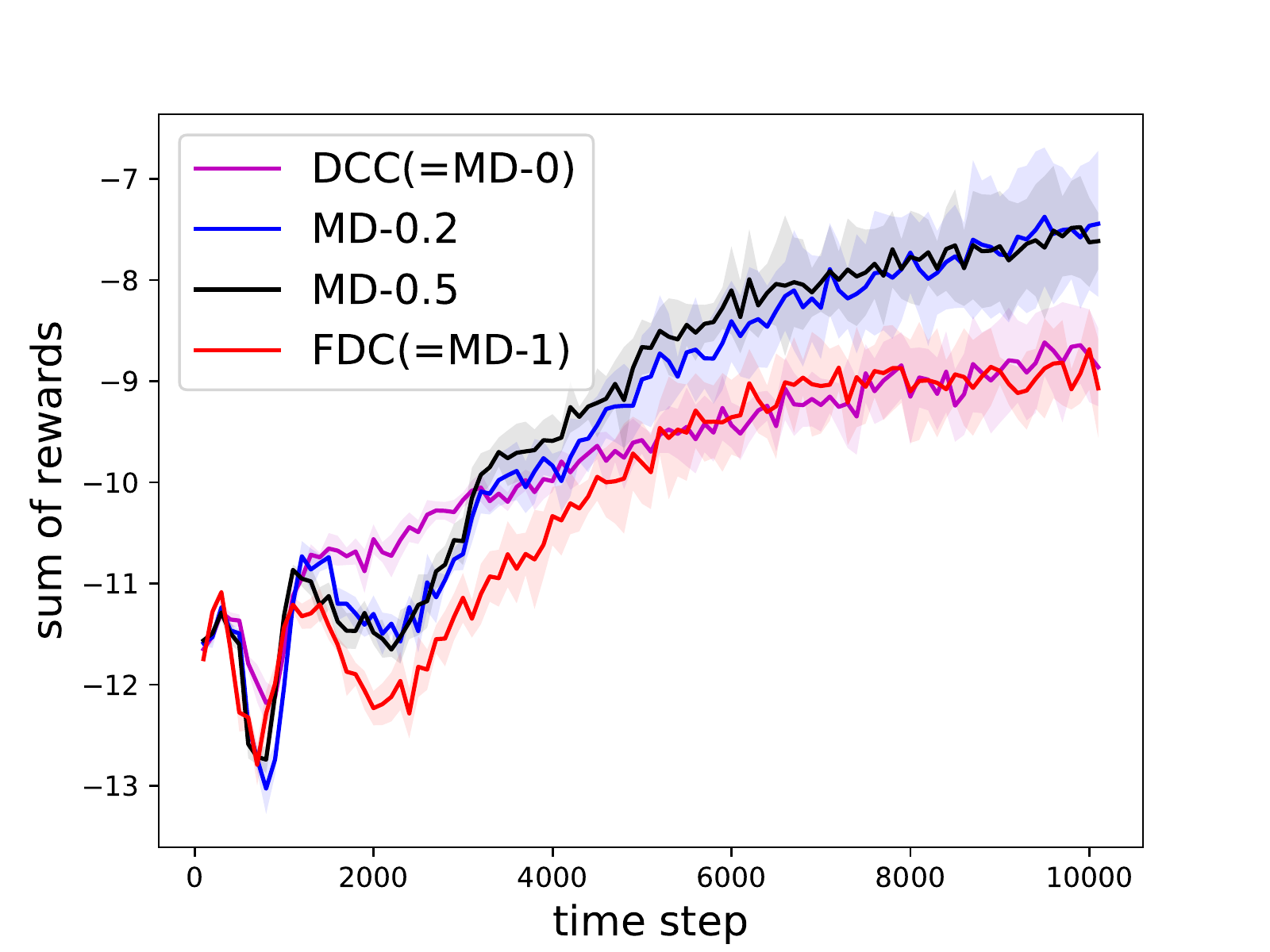}
    \caption{$N=10$}
    \label{fig:res_navi_10}
  \end{subfigure}
  \begin{subfigure}[h]{0.16\textwidth}
    \centering
    \includegraphics[height=2.3cm]{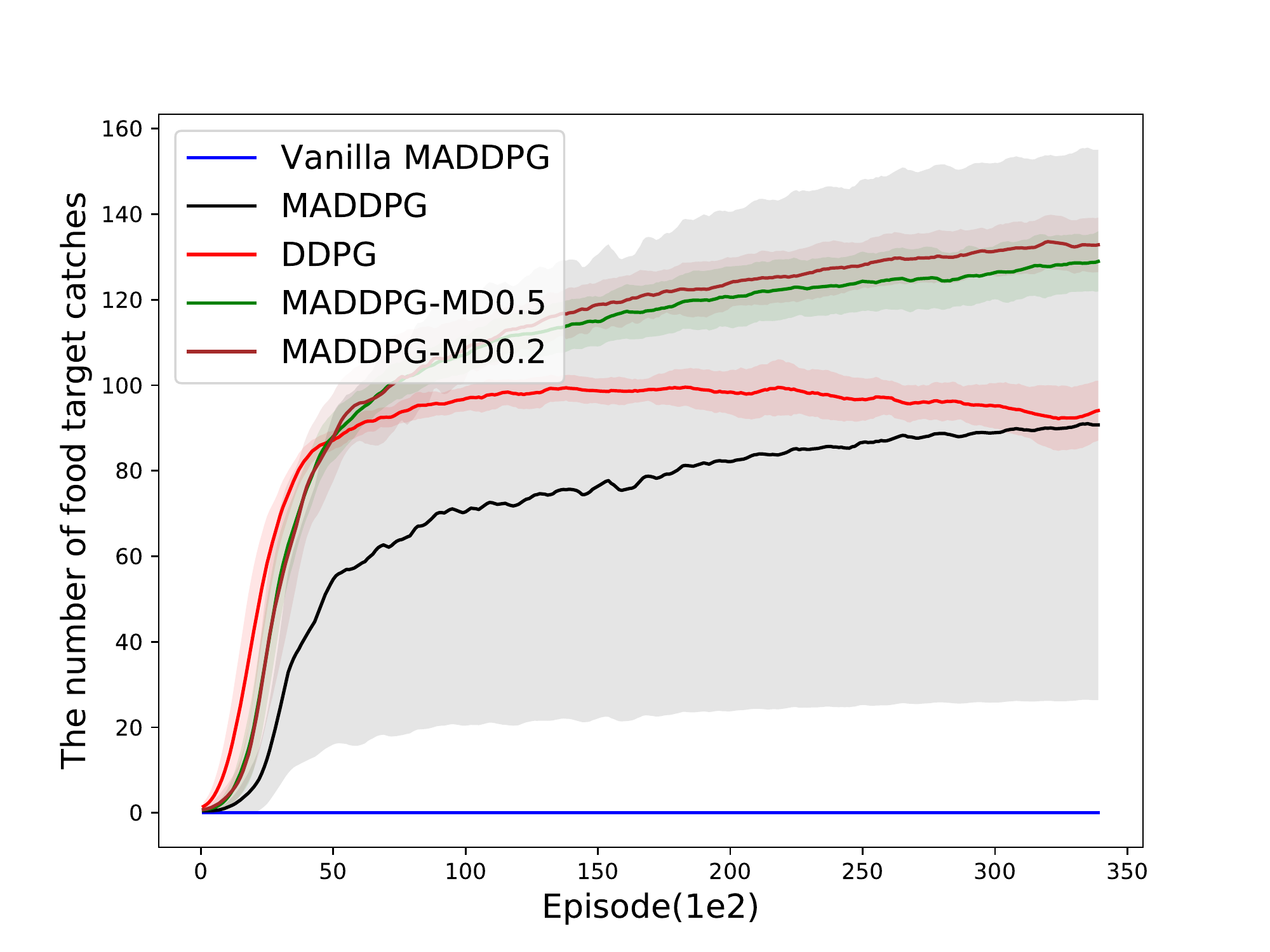}
    \caption{$N=8$}
    \label{fig:res_water_8}
  \end{subfigure}
  \begin{subfigure}[h]{0.16\textwidth}
    \centering
    \includegraphics[height=2.3cm]{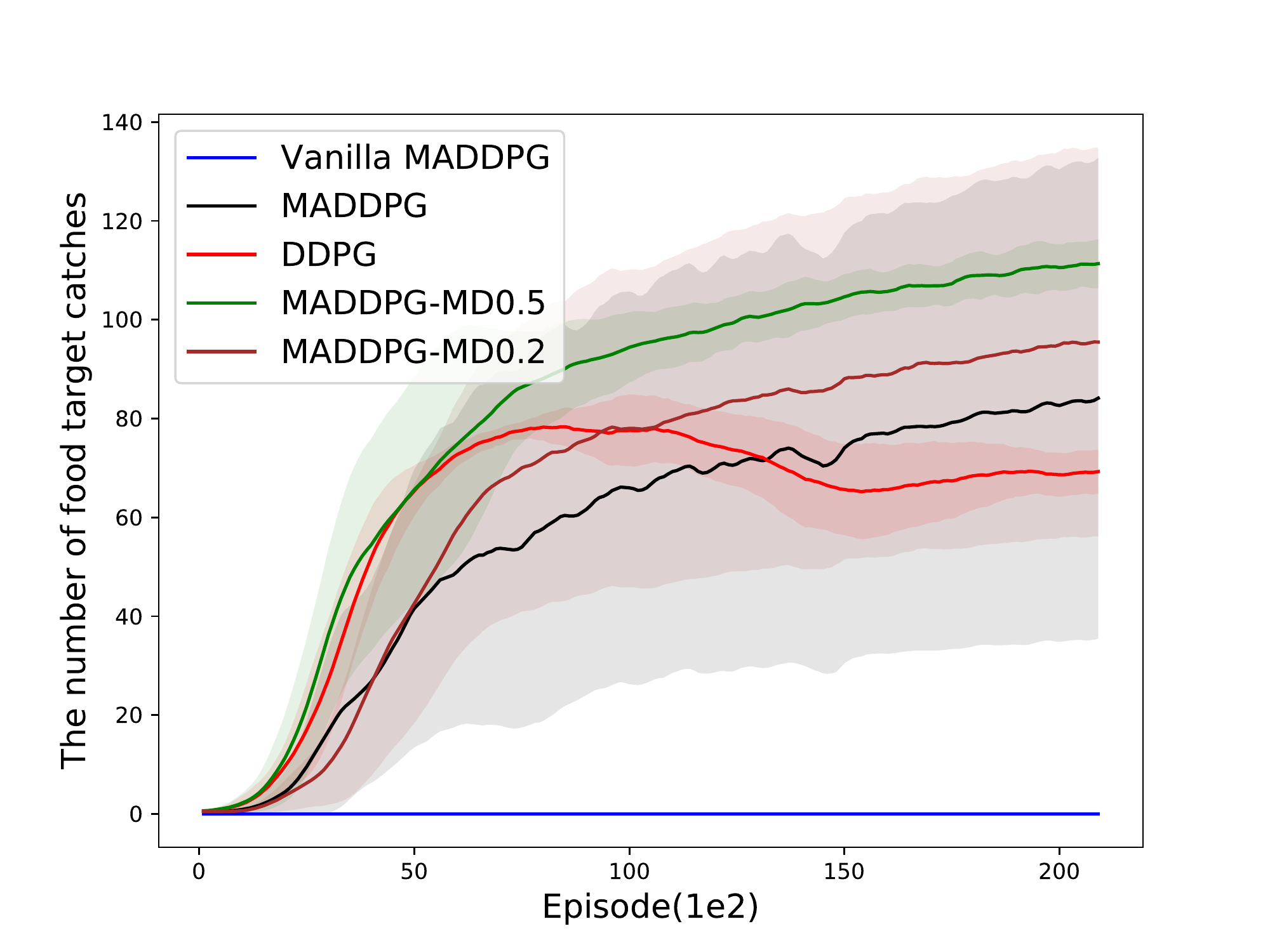}
    \caption{$N=10$}
    \label{fig:res_water_10}
  \end{subfigure}
  \caption{Results - (a)/(b): Pursuit - performance after training as a function of the dropout rate, (c)/(d): cooperative navigation - learning curve, and (e)/(f): waterworld - learning curve)}
\end{figure*}

\subsubsection{Pursuit}

The pursuit game is a standard task for multi-agent systems \cite{Vidal&Shakernia&Kim&Shim&Sastry:02TORA}.
The environment is made up of a two-dimensional grid and consists of  $N$ pursuers and $M$ evaders.
The goal of the game is to capture all evaders as fast as possible by training the agents (i.e., pursuers).
Initially, all the evaders are at the center of the two-dimensional grid, and each evader randomly and independently chooses one of five actions at each time step: move North, East, West,  South, or Stay. (Each evader stays if there exists a pursuer or a map boundary at the position where it is going to move.)
Each pursuer is initially located at a random position of the map and has five possible actions: move North, East, West, South or Stay.
When the four sides of an evader are surrounded by  pursuers or map boundaries, the evader is removed and the pursuers who capture the evader receive  $R^+$ reward.
All pursuers receive $-R_1^-$ reward for each time step and $-R_2^-$ reward if the pursuer hits the  map boundary (the latter negative reward is to promote exploration).
An episode ends when all the evaders are captured or  $T$ time steps elapse.
As in \cite{Gupta&Egorov&Kochenderfer:17AAMAS}, each pursuer observes its surrounding which consists of map boundary, evader(s), or other pursuer(s).
We assume that each pursuer can observe up to $D$ distances in four directions.
Then, the observed information of each pursuer can be represent by a $3\times (2D+1)\times (2D+1)$ observation window (which is the observation of each agent): a $(2D+1)\times (2D+1)$ window  detecting other pursuer(s), a $(2D+1)\times (2D+1)$ window detecting evader(s), and a $(2D+1)\times (2D+1)$ window detecting the map boundary.
For the game of pursuit, we set $R^+=5$, $R_1^-=0.05$, $R_2^-=0.5$, $T=500$, $M=2$ and $D=3$ and simulate two cases: $N=6$ and $N=8$. The map size of the two cases are $15\times 15$ and $17\times 17$ respectively.

\subsubsection{Cooperative navigation}

Cooperative navigation, which was introduced in \cite{mordatch2017emergence}, consists of $N$ agents and $L$ landmarks.
The goal of this environment is to occupy all of the landmarks while avoiding collisions among agents. The observation of each agent is the concatenation of its position and velocity, the locations of landmarks, and  the locations of other agents.
Since we consider partially observable  setting, we assume that each agent observes the locations of other agents only if the distance is less than $D$.
Each agent receives a negative reward $-R_1^{-}$ as the negative of the minimun of the distances from the agent to the $L$ landmarks and receives a negative reward $-R_2^{-}$ if the collision among agents occurs.
In this environment, we set $R_2^{-}=2$, and simulate two cases: $N=8, L=8$ and $N=10, L=10$.

\subsubsection{Waterworld}

Waterworld is an extended environment of pursuit to a continuous domain \cite{Gupta&Egorov&Kochenderfer:17AAMAS}.
The environment is made up of a two-dimensional space and consists of  $N$ pursuers and $M$ food targets, $L$ poison targets, and one obstacle.
The goal of the environment is to capture as many food targets as possible within a given episode of $T$ time steps while avoiding poison targets.
In order to make the game more cooperative, at least $K$ agents need to cooperate to catch a food target.
Each agent takes two-dimensional physical actions to the environment and has observation which consists of its position and information from 25 range-limited sensors of the agent.
The sensors of each agent are used to offer the distances and velocities of other agents, food targets, and poison targets.
The agents receive a reward $R_1^{+}$ when they capture a food target and are penalized by getting reward $-R_1^{-}$ when they encounter a poison target.
To promote exploration, a reward $R_2^{+}$ is given to an agent if the agent touches a food target.
They also receive an action penalty reward defined as the square norm of the action.
In this environment, we set $R_1^{+}=10$, $R_2^{+}=0.01$, $R_1^{-}=0.1$, and $T=500$, and simulate two cases: $N=8, K=4$ and $N=10, K=5$.

The three environments are briefly illustrated in Fig. \ref{fig:env}.

\subsection{Model Architecture}
\label{subsec:model}

Instead of using the concatenation of the agent's own observation and the received messages as the input of the Q-function, we use an architecture for the Q-function that emphasizes the agent's own observation.
The proposed neural network architecture for the Q-function for agent $i$ is represented by
\begin{equation}
Q^i(o^i,a^i,m^i;\theta^i) = h(f(o^i,a^i),g(m^i))
\end{equation}
where $f$ and $g$ are the neural networks that extracts features of own observation and the received messages, respectively, and $h$ is the neural network that produces the expected return by using the output of $f$ and $g$.
In MADDPG, we replace $m^i$ with the concatenation of $o^{-i}$ and $a^{-i}$, where $o^{-i}=(o^1,\cdots,o^{i-1},o^{i+1},\cdots,o^N)$ and $a^{-i}=(a^1,\cdots,a^{i-1},a^{i+1},\cdots,a^N)$.
Note that $f$, $g$ and $h$ are dependent on the task since the input and action dimensions of each task are different.
The detailed structures of $f$, $g$ and $h$ are explained in supplementary material in a separate file.

\subsection{Result}
\label{subsec:Result}

\subsubsection{Pursuit}
\label{subsub:pursuit}

For the pursuit game, we compared DCC-MD with simple DCC and FDC by varying the dropout rate as $p=[0,0.1,0.2,0.3,0.4,0.5,1.0]$.
Note that DCC-MD with $p=0$ corresponds to simple DCC, whereas DCC-MD with $p=1$ corresponds to FDC.
The performance of each algorithm was measured by the number of evader catches in 500 time steps after training. Fig. \ref{fig:res_pur_6} and Fig. \ref{fig:res_pur_8} show the number of evader catches (in 500 time steps after training)  averaged over $1000$ episodes and 7 random seeds, with respect to the drop rate.
It is seen  that the performance improves as the dropout rate increases initially and then the performance deteriorates as the dropout rate further increases after a certain point. In the considered tasks, the best dropout rate is around $[0.2,0.3]$.
It is seen that DCC-MD with proper drop rate significantly outperforms both simple DCC and FDC.
Note that in the case that the number of agents is $N=8$, simple DCC has even worse performance than FDC.
This is because simple DCC does not learn properly due to the large state space for large $N$.

\subsubsection{Cooperative navigation}
\label{subsubsec:navi}

In this environment, we compared DCC-MD with the dropout rate $0.2$ and $0.5$ with simple DCC and FDC.
Figs. \ref{fig:res_navi_8} and \ref{fig:res_navi_10} show the learning curves for  two cases ($N=8, L=8$) and ($N=10, L=10$), respectively. The y-axis is the sum of all agents' rewards averaged over $7$ random seeds, and the x-axis is time step.
It is seen that  DCC-MD with the dropout rate $0.2$ and $0.5$ outperforms simple DCC and FDC.

\subsubsection{Waterworld}
\label{subsubsec:waterworld}

In the waterworld environment, we now considered (independent) DDPG, vanilla MADDPG, MADDPG, and MADDPG-MD with the drop rate
 $0.2$ and $0.5$, and compared their performances. Here, MADDPG is the modified version of vanilla MADDPG to which the proposed network architecture described in Section \ref{subsec:model} is applied.
 Figs. \ref{fig:res_water_8} and \ref{fig:res_water_10} show the learning curves of the four algorithms for  two cases ($N=8, K=4$) and ($N=10, K=5$), respectively. The y-axis is
  the number of food target catches averaged over random seeds, and the x-axis is time step.
  It is seen that MADDPG-MD outperforms both (independent) DDPG and MADDPG. Note that fully decentralized (independent) DDPG has the fastest learning speed at the initial stage due to its small input dimension, but its performance degrades as time step goes because of no cooperation. It is noteworthy that MADDPG-MD almost achieves the initial learning speed of (independent) DDPG while it yields far better performance at the steady state.

\subsection{Ablation Studies}
\label{subsec:sd}

With the verification of the performance gain of the message-dropout technique, we performed in-depth ablation studies regarding the technique with respect to the four key aspects of the technique:
1) the dropout rate, 2) block-wise dropout versus element-wise dropout, 3) retaining agent's own observation without dropout, and 4) the model architecture.

\subsubsection{Dropout rate}

\begin{figure}[t]
  \centering
  \begin{subfigure}[t!]{0.23\textwidth}
    \centering
    \includegraphics[width=\textwidth]{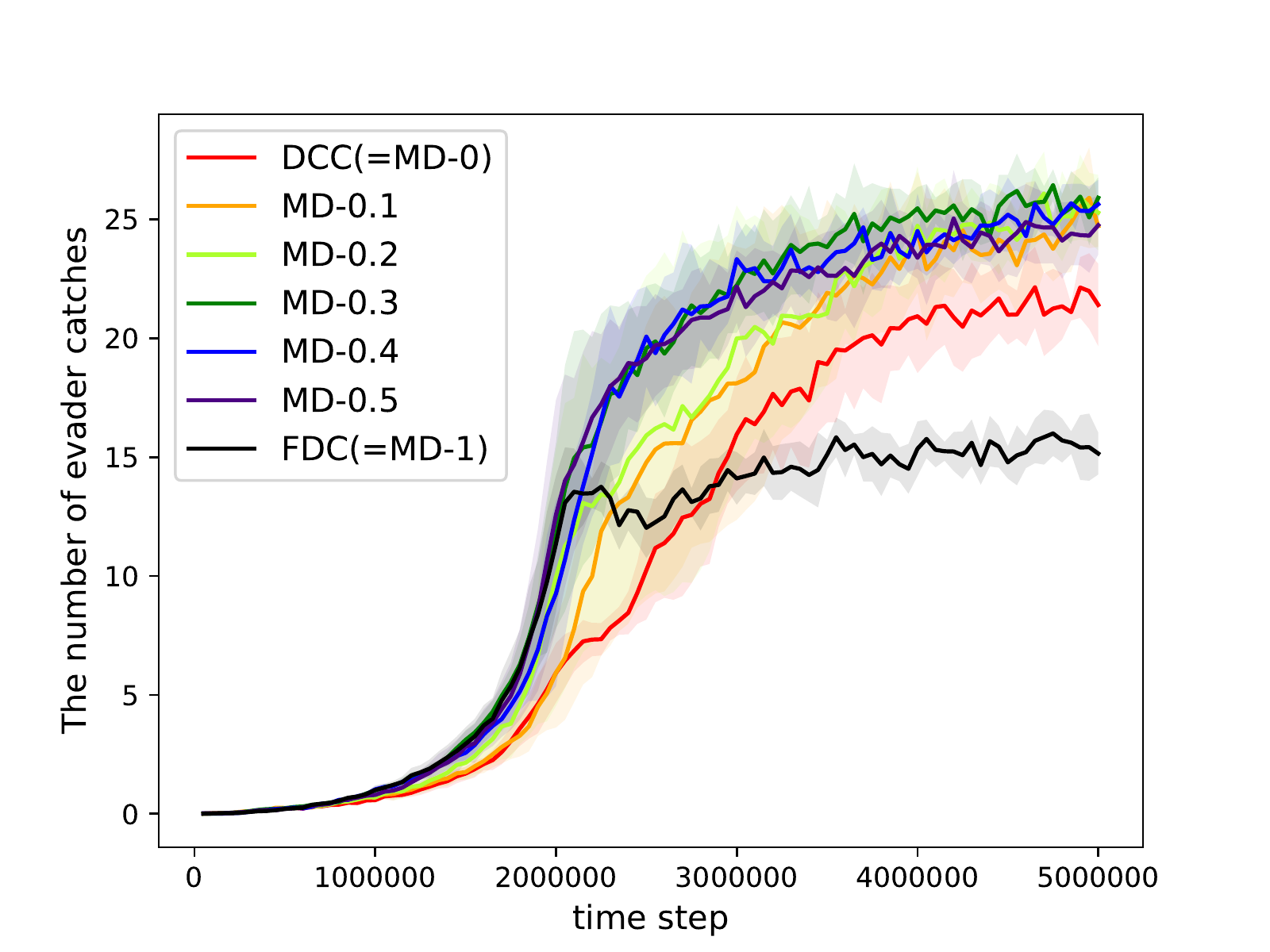}
    \caption{Pursuit, $N=6$}
    \label{fig:res_droprate_6}
  \end{subfigure}
  \begin{subfigure}[t!]{0.23\textwidth}
    \centering
    \includegraphics[width=\textwidth]{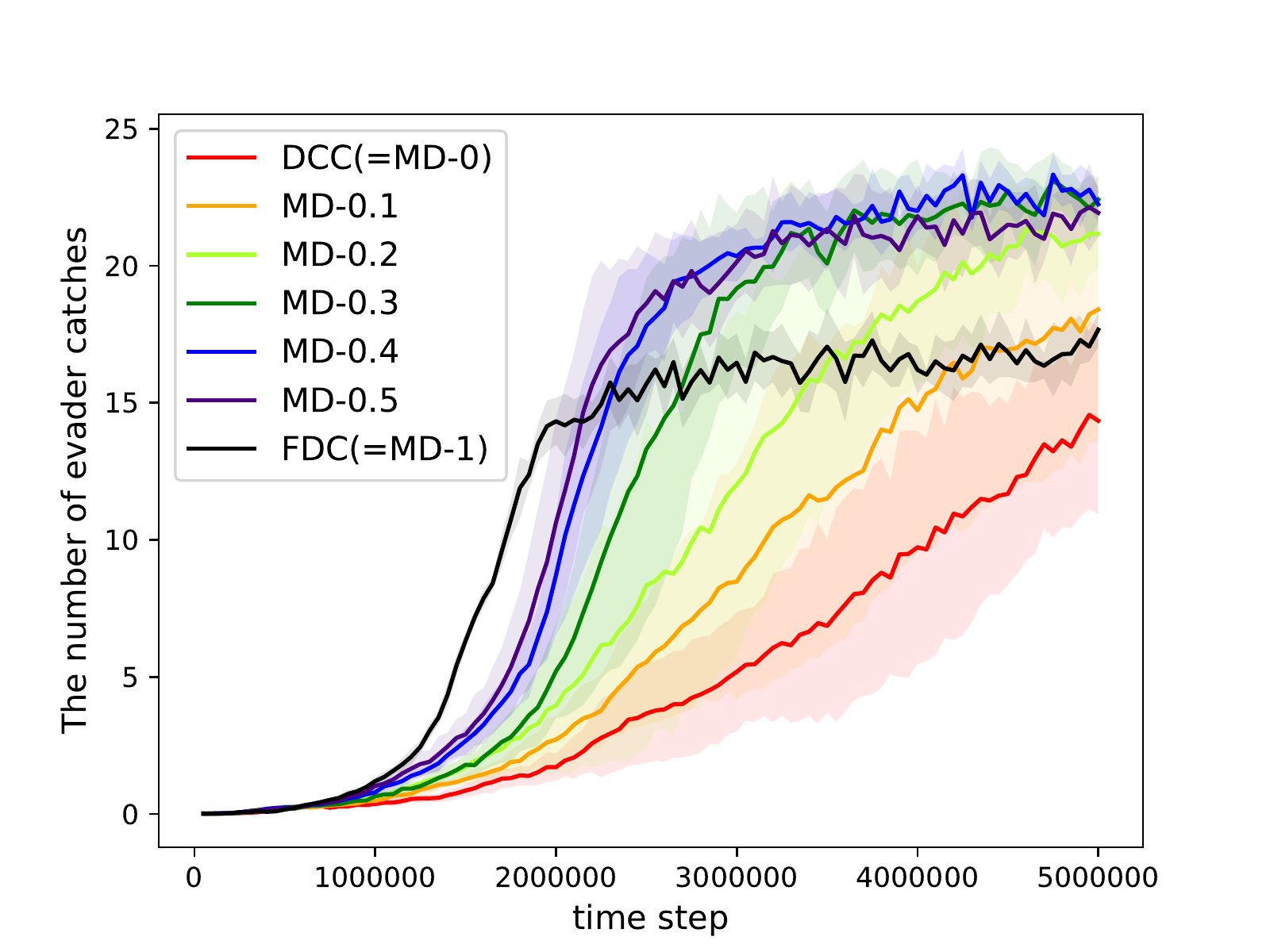}
    \caption{Pursuit, $N=8$}
    \label{fig:res_droprate_8}
  \label{fig:res_droprate}
  \end{subfigure}
  \caption{DCC-MD with respect to dropout rate in pursuit (MD-p: DCC-MD with dropout rate $p$)}
\end{figure}

As mentioned in Section \ref{sec:Background}, we can view that message-dropout generates an ensemble of $2^{N-1}$ differently-thinned networks and averages these thinned networks.
From this perspective, the dropout rate determines the distribution of the thinned networks.
For example, all the $2^{N-1}$ networks are uniformly used to train the ensemble Q-network if the dropout rate is $0.5$.
As the dropout rate increases, the overall input space shrinks in effect and
the portion of the own observation space becomes large in the overall input space, since we apply message-dropout only to the message inputs from other agents.
Hence, it is expected that the learning speed increases especially for large $N$ as the dropout rate increases.
Figs. \ref{fig:res_droprate_6} and \ref{fig:res_droprate} show the learning performance of the algorithms in the training phase. The x-axis is the current time step, and the y-axis is the number of evader catches in $500$ time steps.
As expected, it is seen that  the learning speed increases as the dropout rate increases.
This behavior is clearly seen in Fig. \ref{fig:res_droprate_8}, where the number of agents is $N=8$.
Note that message-dropout with proper drop rate achieves gain in both the learning speed and the steady-state performance.
Figs. \ref{fig:res_pur_6} and \ref{fig:res_pur_8} show the corresponding performance in the execution phase after the training.
It seems that the drop rate of 0.2 to 0.5 yields similar performance and the performance is not so sensitive to the drop rate within this range.

\subsubsection{Block-wise dropout versus element-wise dropout}

We compared message-dropout with standard-dropout which drops the messages of other agents out in an element-wise manner while retaining each agent's own observation.
Fig. \ref{fig:res_sd} shows that message-dropout yields better performance than the standard element-wise dropout.
The difference between message-dropout and standard-dropout is the projected subspaces of the input space.
Message-dropout projects the input space onto $2^{N-1}$ subspaces which always include own observation space, whereas standard-dropout projects the input space onto $2^{(N-1)|\mathcal{M}_i^j|}$ subspaces which contain the projected subspaces by message-dropout.

\subsubsection{Retaining agent's own observation without dropout}

\begin{figure}[t]
  \centering
  \begin{subfigure}[t!]{0.23\textwidth}
    \centering
    \includegraphics[width=\textwidth]{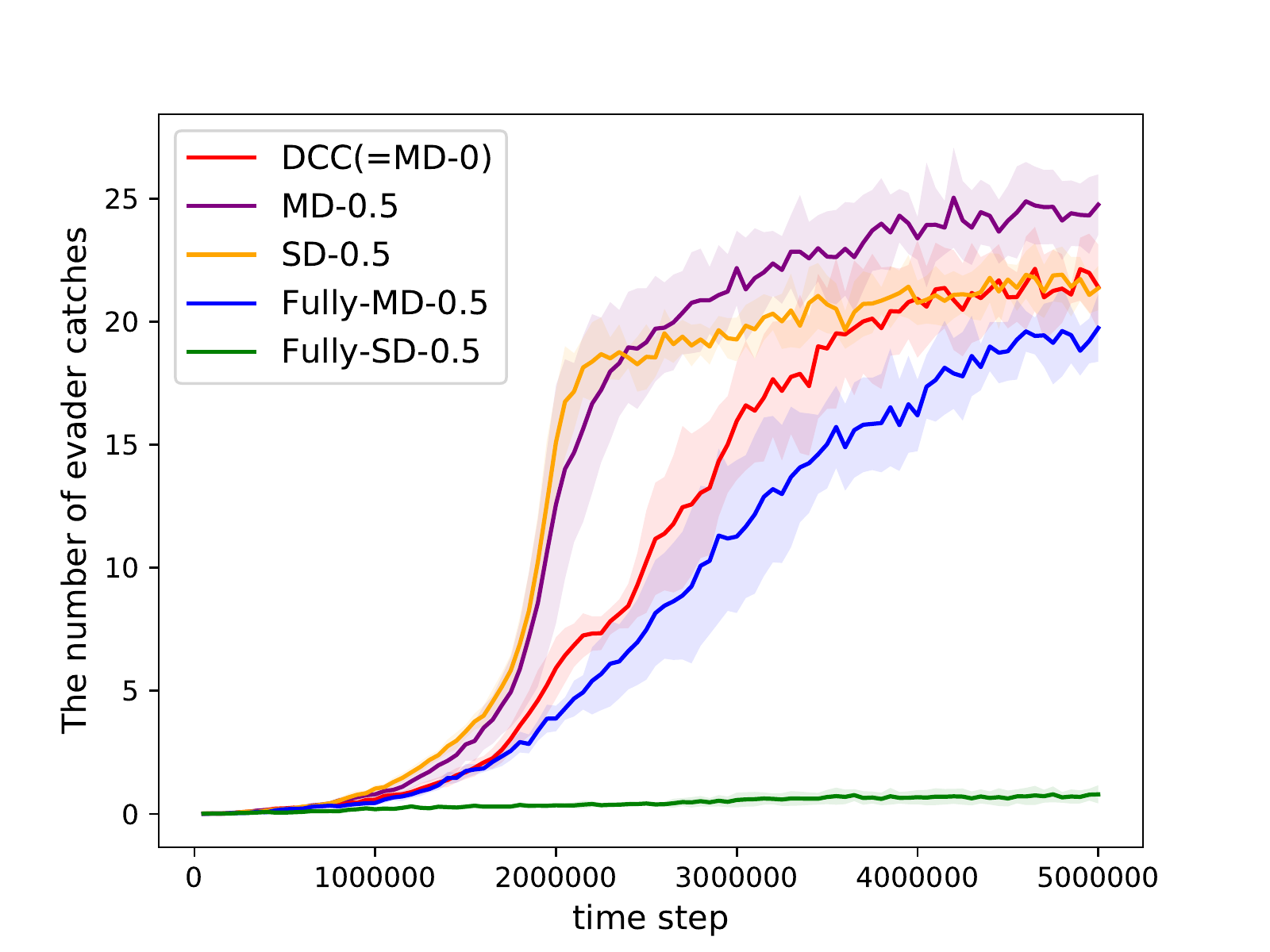}
    \caption{Pursuit, $N=6$}
    \label{fig:res_sd}
  \end{subfigure}
  \begin{subfigure}[t!]{0.23\textwidth}
    \centering
    \includegraphics[width=\textwidth]{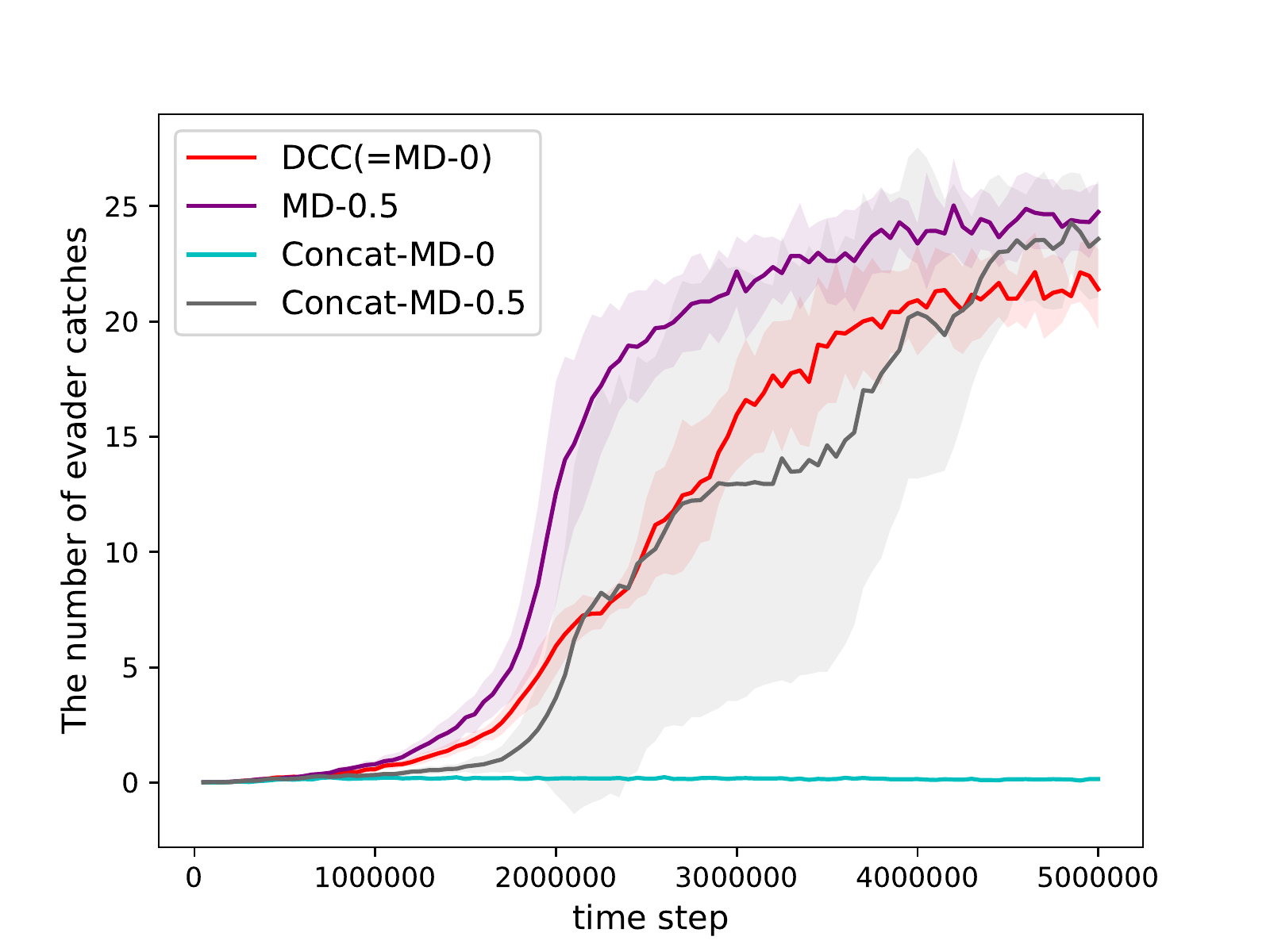}
    \caption{Pursuit, $N=6$}
    \label{fig:res_sd2}
  \end{subfigure}
  \caption{SD-p: standard element-wise dropout with drop rate p, Full MD-p and Full SD-p: dropout includes agent's own observation, Concat-MD-p : DCC-MD-p with the concatenation of own observation and other agents' messages as input}
\end{figure}

We compared message-dropout with full message-dropout which applies message-dropout to each agent's own observation as well as the messages of other agents.
Fig. \ref{fig:res_sd} shows that the full message-dropout increases the training time, similarly to the known fact that in general dropout increases the training time \cite{Srivastava&Hinton&Krizhevsky&Sutskever&Salakhutdinov:14JAIR}.
Whereas full message-dropout yields slower training than simple DCC, message-dropout makes training faster than simple DCC.
Note that standard elementwise dropout without dropping agent's own observation also yields faster training than simple DCC, but full standard elementwise dropout fails to train. Hence, it is important to retain each agent's own observation without dropping out when dropout is applied to MADRL with information exchange.

\subsubsection{Model architecture}

We used the neural network architecture of Q-function that is described in Section \ref{subsec:model} for all environments.
In pursuit with $N=8$ and waterworld, learning failed with
the simple model architecture that uses the concatenation of each agent's own observation and the received messages as input.
Hence, the proposed model architecture is advantageous  when the input space of Q-function is large.
Note that the proposed model architecture has more layers for agent's own observation than those for the received messages from other agents as shown in Fig. \ref{fig:res_sd2}, and hence the feature for more important agent's own observation is well extracted. An interested reader is referred to the supplementary file.

\subsection{Test in The Unstable Environment}
\label{subsec:TestinTheUnstableEnvironment}

Up to now, we have assumed that communication among agents is stable without errors in both training and execution phases. Now,
we consider the situation in which the communication is stable in the training phase but unstable in the actual execution phase.
Such situations occur when the training is done in a controlled environment but the execution is performed in an uncontrolled environment with real deployment.
We considered two communication-unstable cases: case 1 is the case that randomly chosen half of the connections among agents are broken, and case 2 is the case that all connections among agents is broken. When the communication between two agents is broken, we use a zero vector instead of the message received from each other. Note that the performance of FDC does not change since it requires no communication.

\begin{figure}[t]
  \centering
  \begin{subfigure}[t!]{0.42\textwidth}
    \centering
    \includegraphics[width=\textwidth]{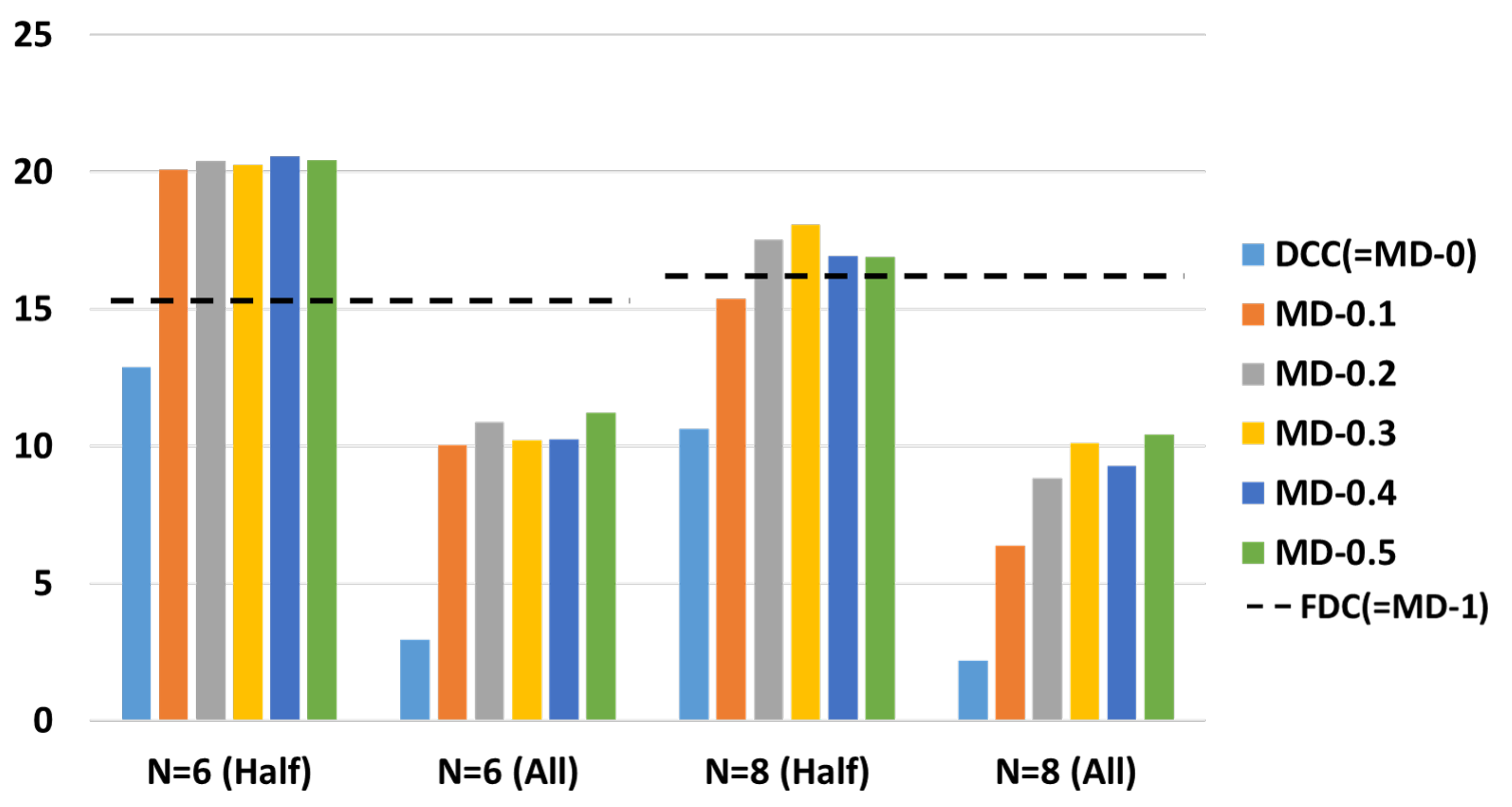}
    \label{fig:result_unstable_1}
  \end{subfigure}
  \caption{(Half) : the performance of each method in the case where the half of connection between agents are broken in pursuit, (All) : The performance of each method in the case where all connection between agents are broken in pursuit}
  \label{fig:result_unstable}
\end{figure}

Fig. \ref{fig:result_unstable} shows the average number of evader catches in the two considered cases in pursuit.
It is seen that DCC-MD outperforms both simple DCC and FDC when the communication link is broken but not all links are broken. It means that message-dropout in the training phase makes the learning robust against communication errors and can still outperform FDC even with other agents' messages coming less frequently.
On the other hand, when the communication link is too unstable, DCC-MD cannot recover this communication loss (but still better than simple DCC), but FDC is better in this case.
Hence, the message-dropout in the training phase can be useful in the situation in which communication among agents is erroneous with a certain probability in the real execution phase.

\section{Related Work}

Recent works in MADRL focus on how to improve the performance compared to FDC composed of independently learning agents.
To harness the benefit from other agents, \cite{Foerster&Whiteson:16NIPS} proposed DIAL, which learns the communication protocol between agents by passing gradients from agent to agent.
\cite{Foerster:17AAAI} proposed the multi-agent actor-critic method called COMA,  which uses a centralized critic to train decentralized actors and a counterfactual baseline to address the multi-agent credit assignment problem.

In most MADRL algorithms such as those mentioned above, the input space of the network (policy, critic, etc.) grows exponentially with the number of agents.
Thus, we expect that message-dropout can be combined with those algorithms to yield better performance.
To handle the increased dimension in MADRL, \cite{Yang:18ICML}  proposed mean field reinforcement learning in which the Q-function is factorized by using the local action interaction
 and approximated by using the mean field theory. Whereas mean field reinforcement learning
 handles the action space within the input space consisting of action and observation, message-dropout can handle not only the action space but also the observation space.

\section{Conclusion}
\label{sec:Conclusion}

In this paper, we have proposed the message-dropout technique for MADRL.
The proposed message-dropout technique effectively handles the increased input dimension in MADRL with information exchange, where each agent uses the information of other agents to train the policy. We have provided ablation studies on the performance of  message-dropout with respect to various aspects of the technique.
The studies show that  message-dropout with proper dropout rates significantly improves performance in terms of the training speed and the steady-state performance. Furthermore, in the scenario of decentralized control with communication, message-dropout makes  learning robust against communication errors in the execution phase.
Although we assume that the communication between agents is fully allowed, message-dropout can be  applied  to the scenario in which communication between limited pairs of agents is available too.

\section{ Acknowledgments}
This work was supported in part by the ICT R$\&$D program of MSIP/IITP (2016-0-00563, Research on Adaptive Machine Learning Technology Development for Intelligent Autonomous Digital Companion) and in part by the National Research Foundation of Korea(NRF) grant funded by the Korea government(Ministry of Science and ICT) (NRF-2017R1E1A1A03070788).

\bibliography{referenceBibs_wjkim_aaai}

\begin{thebibliography}{}

\bibitem[\protect\citeauthoryear{Baldi and
  Sadowski}{2013}]{Baldi&Sadowski:13NIPS}
Baldi, P., and Sadowski, P.~J.
\newblock 2013.
\newblock Understanding dropout.
\newblock In {\em Advances in neural information processing systems},
  2814--2822.

\bibitem[\protect\citeauthoryear{Bu{\c{s}}oniu, Babu{\v{s}}ka, and
  De~Schutter}{2010}]{Busonui&Babuska&Schutter:08IMAS}
Bu{\c{s}}oniu, L.; Babu{\v{s}}ka, R.; and De~Schutter, B.
\newblock 2010.
\newblock Multi-agent reinforcement learning: An overview.
\newblock In {\em Innovations in multi-agent systems and applications-1}.
  Springer.
\newblock  183--221.

\bibitem[\protect\citeauthoryear{Foerster \bgroup et al\mbox.\egroup
  }{2016}]{Foerster&Whiteson:16NIPS}
Foerster, J.; Assael, I.~A.; de~Freitas, N.; and Whiteson, S.
\newblock 2016.
\newblock Learning to communicate with deep multi-agent reinforcement learning.
\newblock In Lee, D.~D.; Sugiyama, M.; Luxburg, U.~V.; Guyon, I.; and Garnett,
  R., eds., {\em Advances in Neural Information Processing Systems 29}. Curran
  Associates, Inc.
\newblock  2137--2145.

\bibitem[\protect\citeauthoryear{Foerster \bgroup et al\mbox.\egroup
  }{2017}]{Foerster&Nardelli&Farquhar&Afouras&Torr&Kohli&Whiteson:17ICML}
Foerster, J.; Nardelli, N.; Farquhar, G.; Afouras, T.; Torr, P. H.~S.; Kohli,
  P.; and Whiteson, S.
\newblock 2017.
\newblock Stabilising experience replay for deep multi-agent reinforcement
  learning.
\newblock In Precup, D., and Teh, Y.~W., eds., {\em Proceedings of the 34th
  International Conference on Machine Learning}, volume~70 of {\em Proceedings
  of Machine Learning Research},  1146--1155.
\newblock International Convention Centre, Sydney, Australia: PMLR.

\bibitem[\protect\citeauthoryear{Foerster \bgroup et al\mbox.\egroup
  }{2018}]{Foerster:17AAAI}
Foerster, J.; Farquhar, G.; Afouras, T.; Nardelli, N.; and Whiteson, S.
\newblock 2018.
\newblock Counterfactual multi-agent policy gradients.

\bibitem[\protect\citeauthoryear{Goldman and
  Zilberstein}{2004}]{Goldman&Zilberstein:04JAIR}
Goldman, C.~V., and Zilberstein, S.
\newblock 2004.
\newblock Decentralized control of cooperative systems: Categorization and
  complexity analysis.
\newblock {\em Journal of artificial intelligence research} 22:143--174.

\bibitem[\protect\citeauthoryear{Gupta, Egorov, and
  Kochenderfer}{2017}]{Gupta&Egorov&Kochenderfer:17AAMAS}
Gupta, J.~K.; Egorov, M.; and Kochenderfer, M.
\newblock 2017.
\newblock Cooperative multi-agent control using deep reinforcement learning.
\newblock In {\em International Conference on Autonomous Agents and Multiagent
  Systems},  66--83.
\newblock Springer.

\bibitem[\protect\citeauthoryear{Hara, Saitoh, and
  Shouno}{2016}]{Hara&Saitoh&Shouno:16ICANN}
Hara, K.; Saitoh, D.; and Shouno, H.
\newblock 2016.
\newblock Analysis of dropout learning regarded as ensemble learning.
\newblock In {\em International Conference on Artificial Neural Networks},
  72--79.
\newblock Springer.

\bibitem[\protect\citeauthoryear{Hausknecht and
  Stone}{2015}]{SDMIA15-Hausknecht}
Hausknecht, M., and Stone, P.
\newblock 2015.
\newblock Deep recurrent q-learning for partially observable mdps.
\newblock In {\em AAAI Fall Symposium on Sequential Decision Making for
  Intelligent Agents (AAAI-SDMIA15)}.

\bibitem[\protect\citeauthoryear{Hinton and
  Salakhutdinov}{2006}]{Hinton&Salakhutdinov:06Science}
Hinton, G.~E., and Salakhutdinov, R.~R.
\newblock 2006.
\newblock Reducing the dimensionality of data with neural networks.
\newblock {\em science} 313(5786):504--507.

\bibitem[\protect\citeauthoryear{Kingma and Ba}{2014}]{Kingma&Ba:15ICLR}
Kingma, D.~P., and Ba, J.
\newblock 2014.
\newblock Adam: A method for stochastic optimization.
\newblock {\em arXiv preprint arXiv:1412.6980}.

\bibitem[\protect\citeauthoryear{Lowe and Mordatch}{2017}]{Lowe&Ryan:17NIPS}
Lowe, Wu, T. H.~A., and Mordatch.
\newblock 2017.
\newblock Multi-agent actor-critic for mixed cooperative-competitive
  environments.
\newblock {\em Advances in Neural Information Processing Systems}.

\bibitem[\protect\citeauthoryear{Mordatch and
  Abbeel}{2017}]{mordatch2017emergence}
Mordatch, I., and Abbeel, P.
\newblock 2017.
\newblock Emergence of grounded compositional language in multi-agent
  populations.
\newblock {\em arXiv preprint arXiv:1703.04908}.

\bibitem[\protect\citeauthoryear{Srivastava \bgroup et al\mbox.\egroup
  }{2014}]{Srivastava&Hinton&Krizhevsky&Sutskever&Salakhutdinov:14JAIR}
Srivastava, N.; Hinton, G.; Krizhevsky, A.; Sutskever, I.; and Salakhutdinov,
  R.
\newblock 2014.
\newblock Dropout: A simple way to prevent neural networks from overfitting.
\newblock {\em The Journal of Machine Learning Research} 15(1):1929--1958.

\bibitem[\protect\citeauthoryear{Tampuu \bgroup et al\mbox.\egroup
  }{2017}]{Tampuu&Matiisen&Kodelja&Kuzovkin&Korjus&Aru&Vicente:17PLOS}
Tampuu, A.; Matiisen, T.; Kodelja, D.; Kuzovkin, I.; Korjus, K.; Aru, J.; Aru,
  J.; and Vicente, R.
\newblock 2017.
\newblock Multiagent cooperation and competition with deep reinforcement
  learning.
\newblock {\em PloS one} 12(4):e0172395.

\bibitem[\protect\citeauthoryear{Tan}{1993}]{Tan&Ming:93ICML}
Tan, M.
\newblock 1993.
\newblock Multi-agent reinforcement learning: Independent vs. cooperative
  agents.
\newblock In {\em Proceedings of the tenth international conference on machine
  learning},  330--337.

\bibitem[\protect\citeauthoryear{Van~Hasselt, Guez, and
  Silver}{2016}]{Hasselt&Guez&Silver:16AAAI}
Van~Hasselt, H.; Guez, A.; and Silver, D.
\newblock 2016.
\newblock Deep reinforcement learning with double q-learning.
\newblock In {\em AAAI}, volume~16,  2094--2100.

\bibitem[\protect\citeauthoryear{Vidal \bgroup et al\mbox.\egroup
  }{2002}]{Vidal&Shakernia&Kim&Shim&Sastry:02TORA}
Vidal, R.; Shakernia, O.; Kim, H.~J.; Shim, D.~H.; and Sastry, S.
\newblock 2002.
\newblock Probabilistic pursuit-evasion games: theory, implementation, and
  experimental evaluation.
\newblock {\em IEEE transactions on robotics and automation} 18(5):662--669.

\bibitem[\protect\citeauthoryear{Yang \bgroup et al\mbox.\egroup
  }{2018}]{Yang:18ICML}
Yang, Y.; Luo, R.; Li, M.; Zhou, M.; Zhang, W.; and Wang, J.
\newblock 2018.
\newblock Mean field multi-agent reinforcement learning.
\newblock In Dy, J., and Krause, A., eds., {\em Proceedings of the 35th
  International Conference on Machine Learning}, volume~80 of {\em Proceedings
  of Machine Learning Research},  5567--5576.
\newblock Stockholmsmässan, Stockholm Sweden: PMLR.

\end{thebibliography}
\bibliographystyle{aaai}

\section{Supplementary Material}

\subsection{Model Architecture and Training Detail}

In this supplementary material, we describe the detailed structure of Q-function for DCC  and MADDPG  and the procedure of training for each task. The Q-function neural network structure for DCC-MD is the same as that of DCC, while only block-dropout is applied in the case of DCC-MD. Similarly, the Q-function neural network structure for MADDPG-MD is the same as that of MADDPG while only block-dropout is applied in the case of DCC-MD.
As mentioned in the main paper, our neural network architecture of Q-function can be expressed as
\begin{equation}
Q^i(o^i,a^i,m^i;\theta^i) = h(f(o^i,a^i),g(m^i)),
\end{equation}
where $f$, $g$, and $h$ are the sub-neural-networks designed properly depending on each task.

\paragraph{Pursuit}

\begin{figure*}[h]
  \centering
  \begin{subfigure}[h]{0.8\textwidth}
    \centering
    \includegraphics[width=\textwidth]{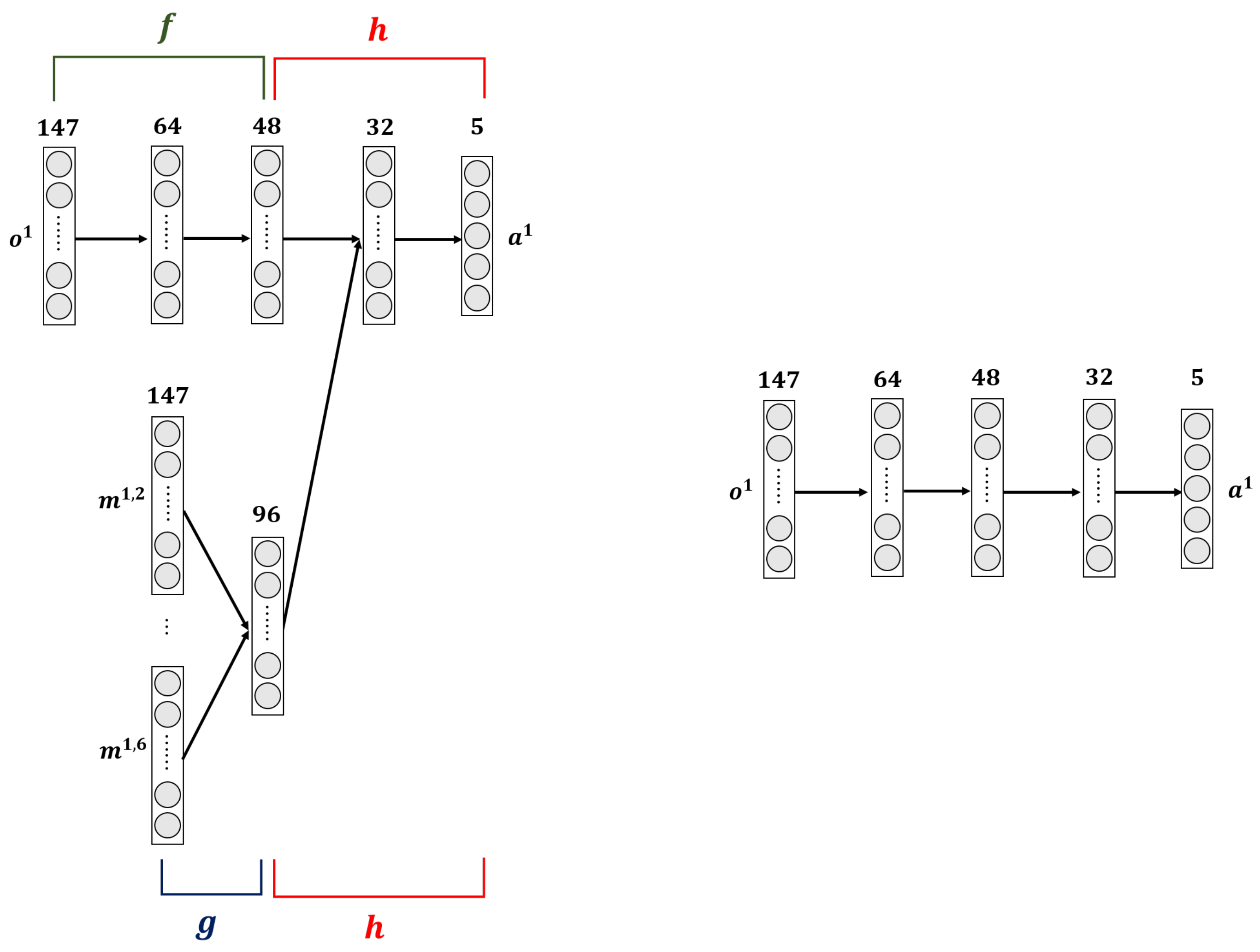}
  \end{subfigure}
  \caption{Agent 1's Q-function neural network architecture for the  pursuit game  with the number of agents $N=6$: (Left) DCC or  DCC-MD and (Right) FDC}
  \label{fig:model_architecture}
\end{figure*}

Fig. \ref{fig:model_architecture} shows the Q-function neural network architecture for DCC (or DCC-MD) and FDC used in the pursuit game with the number of agents  $N=6$.

In pursuit, the observation of three 2D windows $3 \times (2D+1) \times (2D+1)=3\times 7\times 7$ is flattened as $147$ input units. $f$ is a two multi-layer perceptron (MLP) with 64 hidden units and produces a 48-dimensional output.
$g$ is a single-layer perceptron and produces a 96-dimensional output.
Since the action is discrete in pursuit, $a^i$ is not set as an input of $f$ but as the output of $h$, and
$h$ is a two MLP with 32 hidden units.
The activation function of $f$, $g$ and $h$ are ReLU expect the final linear layer of $h$.
In FDC, i.e., the fully decentralized DDQN, each agent has 4 MLP whose activation functions are ReLU except the final linear layer. The four layers have 64, 48, 32, and 5 units, respectively.
In the case of $N=8$, only $g$ is changed to a single-layer perceptron producing a 128-dimensional output

All algorithms, FDC, DCC, and DCC-MD, used the $\epsilon$-greedy policy with $\epsilon$ annealed from 1 to 0.02 over initial $2\times 10^6$ time steps and fixed at 0.02 thereafter. Although it is known that experience replay harms the performance in MADRL in general \cite{Foerster&Nardelli&Farquhar&Afouras&Torr&Kohli&Whiteson:17ICML}, we used experience replay because we observed performance improvement with experience replay for our tasks. Each agent used the replay memory size of $2\times 10^5$. We used $\lambda = 0.99$ and Adam optimizer \cite{Kingma&Ba:15ICLR} with the learning rate $10^{-4}$. For all tasks, we updated the Q-function every 4 time steps.

\paragraph{Cooperative Navigation}

The basic architecture of the Q-function neural network used for cooperative navigation is the same as that shown in Fig. \ref{fig:model_architecture}, except the number of nodes in each layer.

The dimension of observation for cooperative navigation  is $4\times N+2$. $f$ is two MLP with 64 hidden units and produces a 64-dimensional output. $g$ is a single-layer perceptron and produces a 64-dimensional output. $h$ is 2 MLPs with 32 hidden units. The activation function of $f$, $g$ and $h$ is ReLU expect the final linear layer of $h$.
In FDC, i.e., the fully decentralized DDQN, each agent has 4 MLP whose activation functions are ReLU except the final linear layer. The four layers have 64, 64, 32, 5 units, respectively.

All algorithms (FDC, DCC, and DCC-MD) used the $\epsilon$-greedy policy with $\epsilon$ annealed from 1 to 0.02 over $4\times 10^5$ time steps and fixed at 0.02 thereafter. Each agent used the replay memory size of $2\times 10^5$. We used $\lambda = 0.99$ and Adam optimizer with the learning rate $10^{-4}$. For all tasks, we updated the Q-function every 4 time steps.

\paragraph{Waterworld}

\begin{figure*}[h]
  \centering
  \begin{subfigure}[h]{0.8\textwidth}
    \centering
    \includegraphics[width=\textwidth]{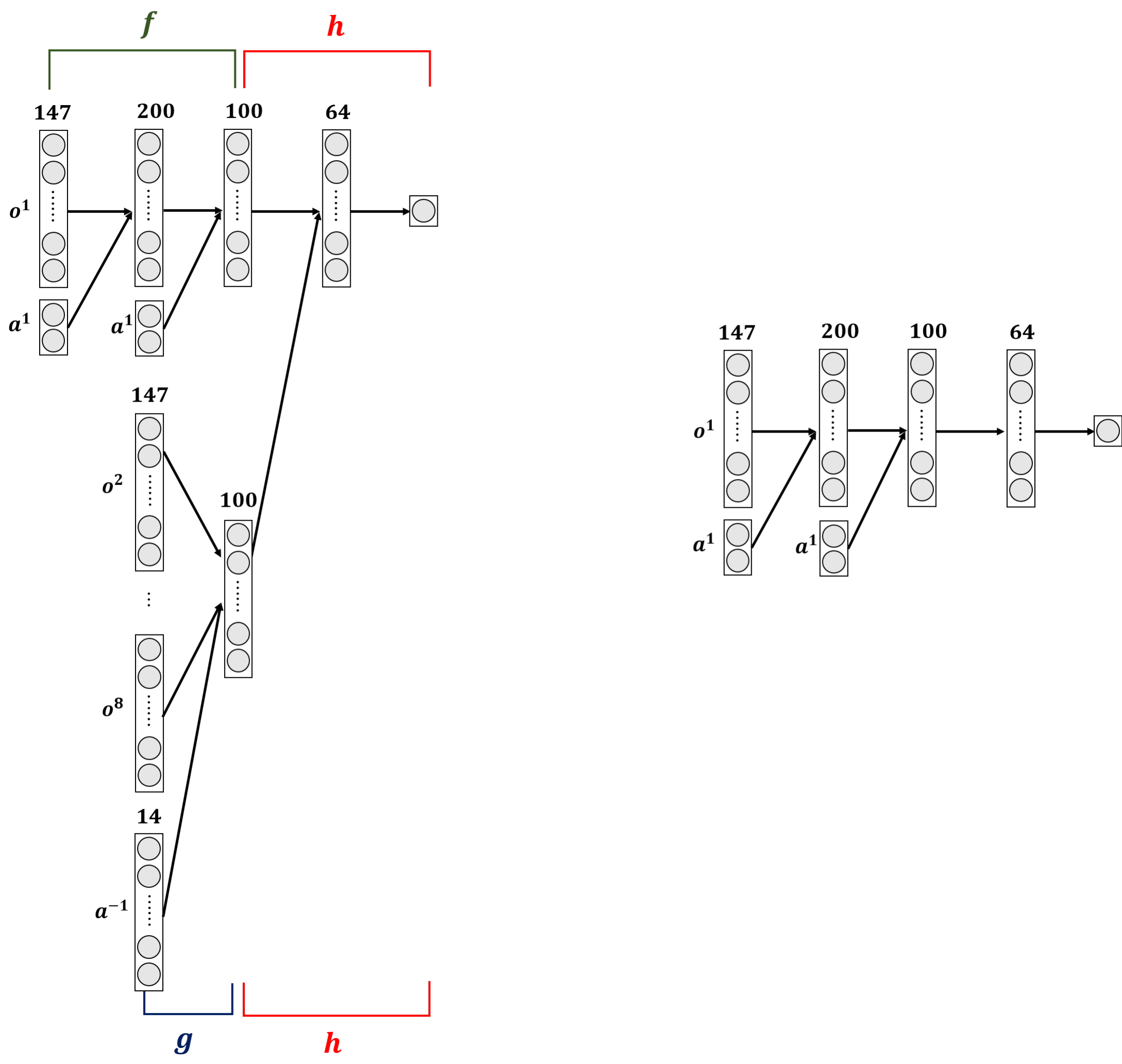}
  \end{subfigure}
  \caption{Agent 1's Q-function neural network architecture for the waterworld game with the number of agents $N=8$: (Left) MADDPG or MADDPG-MD and (Right) DDPG}
  \label{fig:model_arch_waterworld}
\end{figure*}

Fig.  \ref{fig:model_arch_waterworld} shows the neural network architecture used for the waterworld game. MADDPG and MADDPG-MD have the same neural network architecture, while block-dropout is applied to the message input units from other agents in the case of MADDPG-MD.

The neural network architecture of the decentralized actor in MADDPG is two MLP with 64 hidden units.
The neural network architecture  of the centralized critic is expressed  by $f$, $g$ and $h$ as mentioned previously.
 $f$ is 2 MLP with 200 hidden units and produces a 100-dimensional output. Here, the action is included twice to the input layer and the hidden layer in $f$. $g$ is a single-layer perceptron and produces 100 hidden units. $h$ is 2 MLP with 64 hidden units. The activation function of $f$, $g$ and $h$ are ReLU except the final linear layer of $h$.

In the environment of waterworld, all agents share the parameter of critic and actors to promote learning. Each agent used the replay memory size of $5\times10^5$ and Gaussian noise process with $\sigma=0.15$ for efficient exploration. We used $\gamma=0.95$ and Adam optimizer with the learning rate $10^{-3}$. We updated the centralized critic every 5 time steps and the decentralized actors every 10 time steps.

\subsection{Compressing Message using Autoencoder}

\begin{figure}[h]
  \centering
  \begin{subfigure}[h]{0.5\textwidth}
    \centering
    \includegraphics[width=\textwidth]{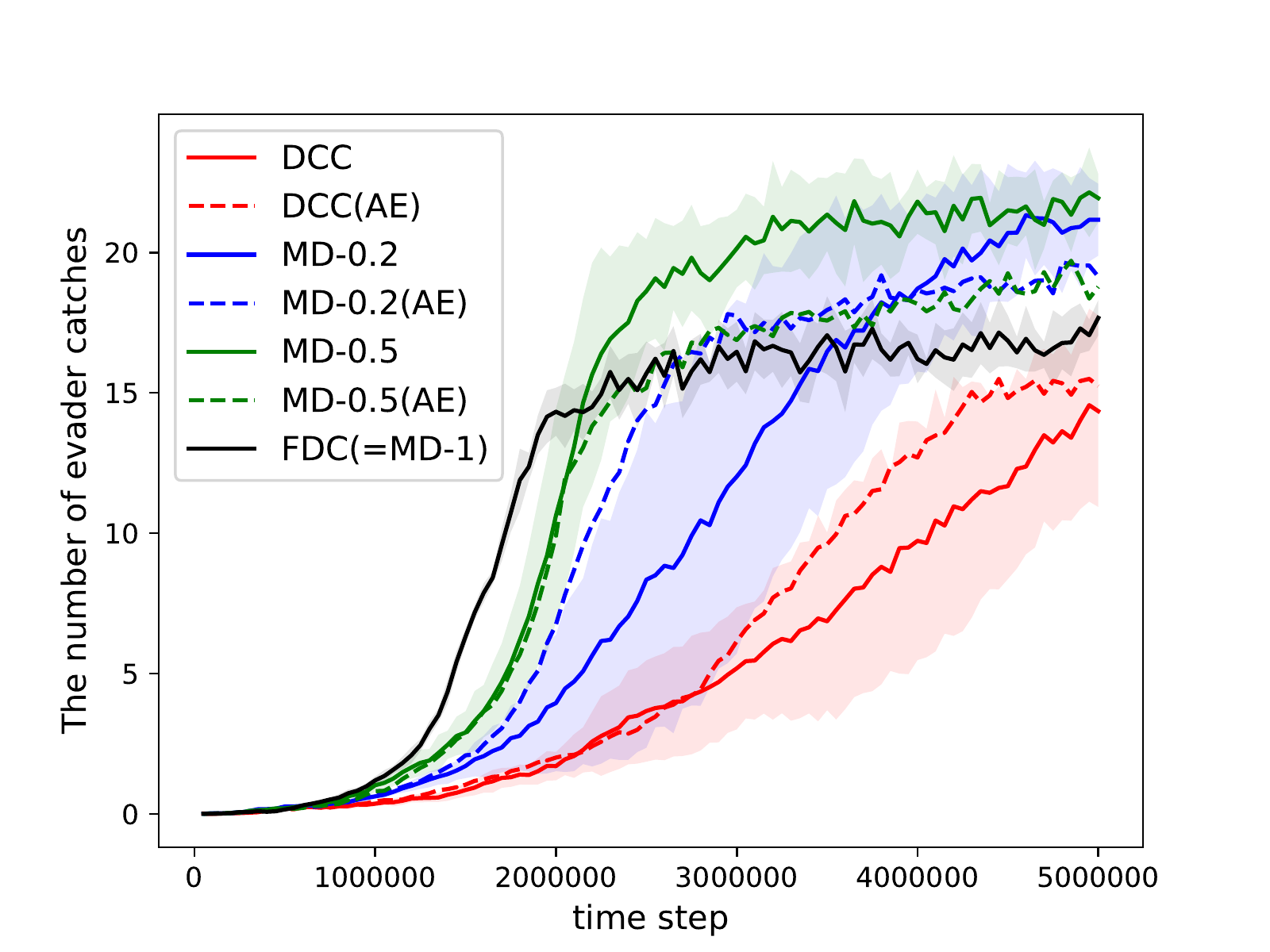}
  \end{subfigure}
  \caption{Pursuit, N=8}
  \label{fig:ae}
\end{figure}

For the purpose of the experiment, we have assumed that the messages are the observation of other agents in the main paper. Message-dropout can also be applied in the scenario of DCC with the more practical assumption that communication load is limited. We compress each agent's observation using the traditional autoencoder introduced in \cite{Hinton&Salakhutdinov:06Science}, and then use it as a message. In the environment of pursuit (N=8), we consider the simple DCC and DCC-MD which are applied autoencoder. Fig. \ref{fig:ae} shows that the learning speed increases in the case of simple DCC and DCC-MD with dropout rate 0.2. Note that DCC-MD still performs better than simple DCC when the messages are compressed version of observation.
However, the compressed messages using autoencoder degrades the steady-state performance.

\begin{figure*}[h]
  \centering
  \begin{subfigure}[h]{0.35\textwidth}
    \centering
    \includegraphics[width=\textwidth]{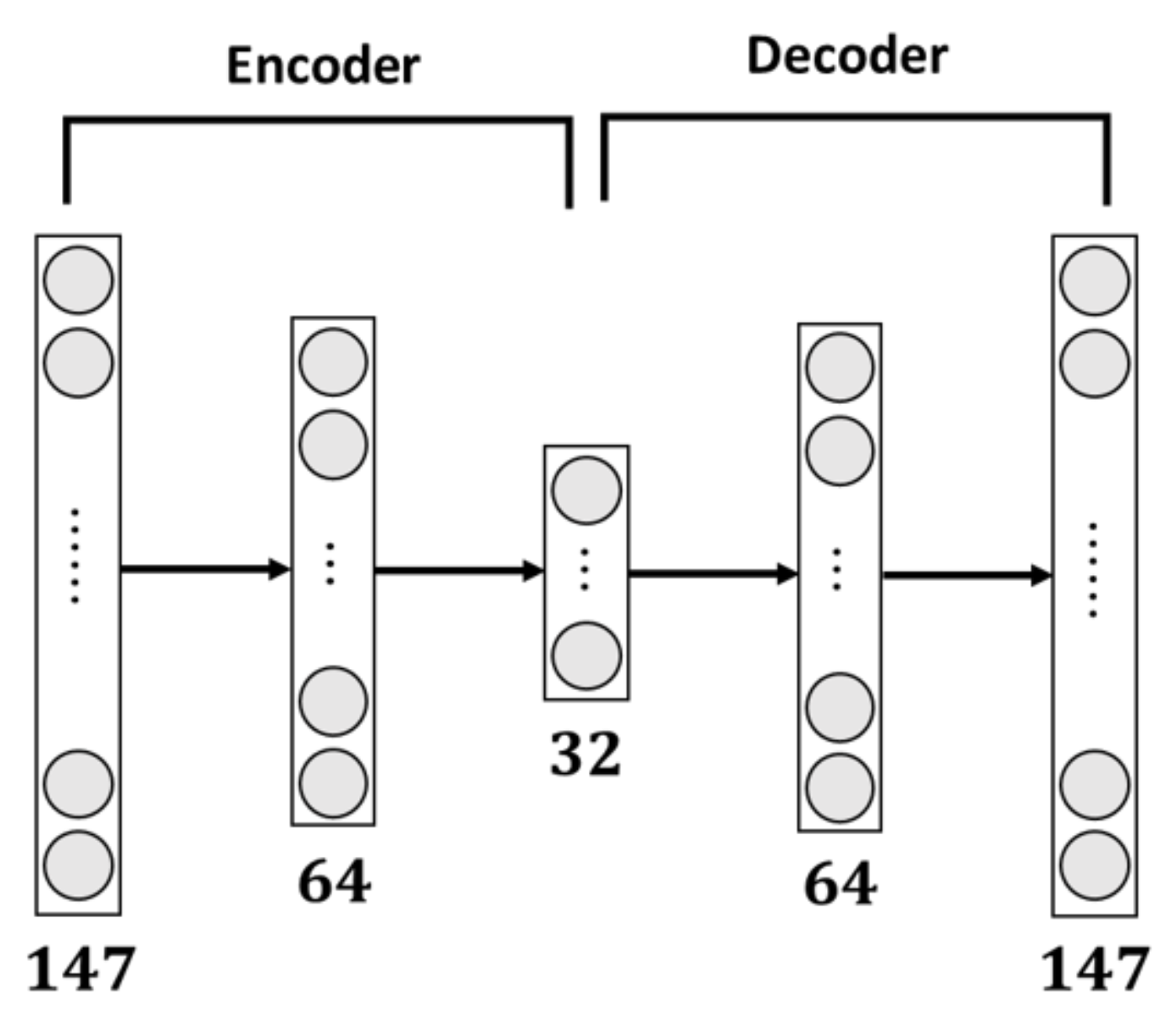}
  \end{subfigure}
  \caption{The neural network architecture for autoencoder in the pursuit game}
  \label{fig:model_ae}
\end{figure*}

Fig. \ref{fig:model_ae} shows the autoencoder architecture used in the pursuit game with the number of agents  $N=8$.
As shown in fig. \ref{fig:model_ae}, the autoencoder consists of 2-MLP encoder and 2-MLP decoder, and 147-dimensional observation is compressed to 32-dimensional message. We pretrained the autoencoder using $10^6$ samples and then used the encoder to compress the observation in training.

\end{document}